\pdfoutput=1

\documentclass[10pt,twocolumn,letterpaper]{article}

\usepackage[pagenumbers]{cvpr} %

\usepackage[dvipsnames]{xcolor}

\usepackage{booktabs}
\usepackage{makecell}
\usepackage{multirow}

\definecolor{cvprblue}{rgb}{0.21,0.49,0.74}
\definecolor{darkgreen}{rgb}{0,0.5,0}
\definecolor{darkblue}{rgb}{0,0.,0.5}
\definecolor{fullred}{rgb}{0.95,.0,.1}
\definecolor{brown}{rgb}{0.65,0.16,0.16}
\definecolor{orange}{rgb}{1,0.5,0}
\newcommand{\comment}[3]{\ifthenelse{\boolean{}}
{2
  \marginpar{\tiny\noindent{\raggedright{{\colorbox{#3}{\sffamily\textcolor{white}{#1
            [\arabic{0}]}}}}} \color{#3}{#2} \par}}{}}

\def\shownotes{1} 
 \ifnum\shownotes=1
\newcommand{\authnote}[2]{{[#1: #2]}}
\else 
\newcommand{\authnote}[2]{{}}
\fi

\newcommand{\xseq}[1]{$\{x^{1:N}_{t}\}_{t=0}^T$}
\newcommand{\dynamic}[1]{\ensuremath{T}}
\newcommand{\name}[1]{HDP}
\newcommand{\fname}[1]{Hierarchical Diffusion Policy}
\newcommand{\cdp}[1]{RK-Diffuser}
\newcommand{\joint}{\mathrm{joint}}
\newcommand{\pose}{\mathrm{pose}}

\colorlet{shadecolor}{gray!15}

\newcommand{\vv}[1]{\boldsymbol{#1}}

\newcommand{\keep}[1]{}
\newcommand{\old}[1]{}

\usepackage[pagebackref,breaklinks,colorlinks,citecolor=cvprblue]{hyperref}

\def\paperID{13857} %
\def\confName{CVPR}
\def\confYear{2024}

\title{Hierarchical Diffusion Policy\\ for Kinematics-Aware Multi-Task Robotic Manipulation}

\author{Xiao Ma, Sumit Patidar, Iain Haughton, Stephen James\\
Dyson Robot Learning Lab\\
{\tt\small \{xiao.ma, sumit.patidar, iain.haughton, stephen.james\}@dyson.com}
}

\begin{document}
\maketitle
\begin{abstract}
This paper introduces \fname{} (\name{}), a hierarchical agent for multi-task robotic manipulation. \name{} factorises a manipulation policy into a hierarchical structure: a high-level task-planning agent which predicts a distant next-best end-effector pose (NBP), and a low-level goal-conditioned diffusion policy which generates optimal motion trajectories. The factorised policy representation allows \name{} to tackle both long-horizon task planning while generating fine-grained low-level actions. To generate context-aware motion trajectories while satisfying robot kinematics constraints, we present a novel kinematics-aware goal-conditioned control agent, Robot Kinematics Diffuser (\cdp{}). Specifically, \cdp{} learns to generate both the end-effector pose and joint position trajectories, and distill the accurate but kinematics-unaware end-effector pose diffuser to the kinematics-aware but less accurate joint position diffuser via differentiable kinematics.  Empirically, we show that \name{} achieves a significantly higher success rate than the state-of-the-art methods in both simulation and real-world.\footnote{Code and videos are available in our \href{https://yusufma03.github.io/projects/hdp/}{project page}.}
\end{abstract}

\section{Introduction}

\label{sec:intro}

Learning efficient visual manipulation strategies in robotics is challenging due to diverse environments, objects, and robot trajectories. The choice of policy representation strongly influences agent performance.

One way of parameterising the policy is to directly map visual observations to robot commands, e.g., joint position or velocity actions~\cite{james2017transferring, matas2018sim, kalashnikov2018qt,wu2019learning}. These approaches make the least assumptions of the task and environment and retain the flexible control of the over-actuated, but they often suffer from low sample efficiency and poor generalisation ability, especially for long-horizon tasks~\cite{james2022coarse,shridhar2023perceiver}.

\begin{figure}
    \centering
    \includegraphics[width=\linewidth]{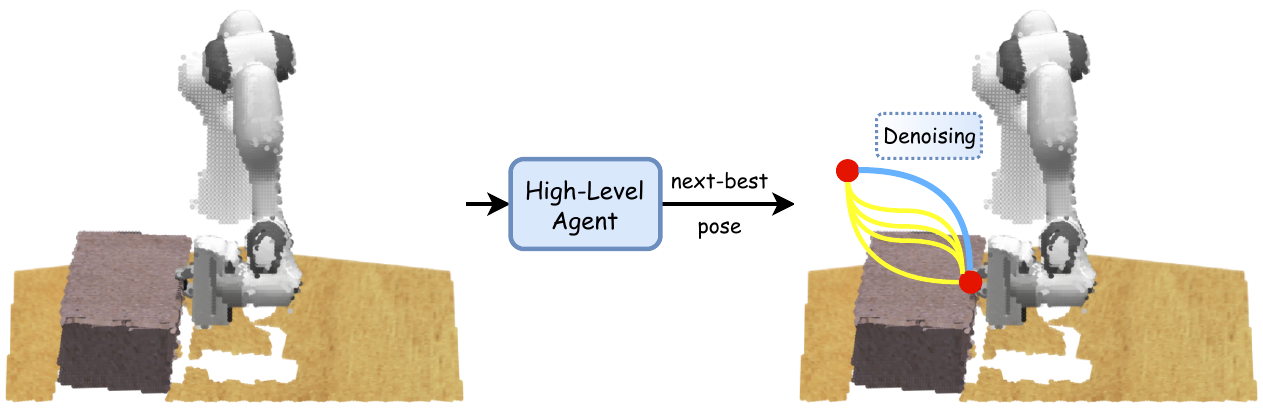}
    \caption{We introduce \name{}, a hierarchical agent for robotic manipulation. At the high-level, \name{} learns to predict the next-best end-effector pose. Conditioned on the current and the predicted pose (\textcolor{red}{red}), a diffusion model generates an action trajectory for the robot to follow (\textcolor{blue}{blue}). In contrast, the trajectories generated by classic planners (\textcolor{yellow}{yellow}) cannot be executed due to violating environment constraints, e.g., the hinge of the box.}\vspace{-5mm}
    \label{fig:teaser}
\end{figure}

Recent advances in learning next-best-pose (NBP) agents~\cite{james2022q, james2022coarse, james2022lpr, james2022tree,zhao2022effectiveness,shridhar2023perceiver,goyal2023rvt,gervet2023act3d} have significantly improved the sample efficiency and performance for robotic manipulation. 
Instead of learning continuous actions, NBP agents directly predict a distant ``keyframe"~\cite{james2022q}, a next-best end-effector pose, and use a predefined motion planner to compute a trajectory for the agent to follow. However, as the motion planner is unaware of the task context, it will fail to perform tasks that require understanding the environment context, e.g., dynamics. For example, in Fig.~\ref{fig:teaser} to open the box, the agent has to understand the unknown physics properties of the hinge, e.g., the resistance force, and only a specific curved trajectory can be successfully executed.

\begin{figure*}
	\centering
	\begin{tabular}{cccc}
    \includegraphics[width=0.23\linewidth]{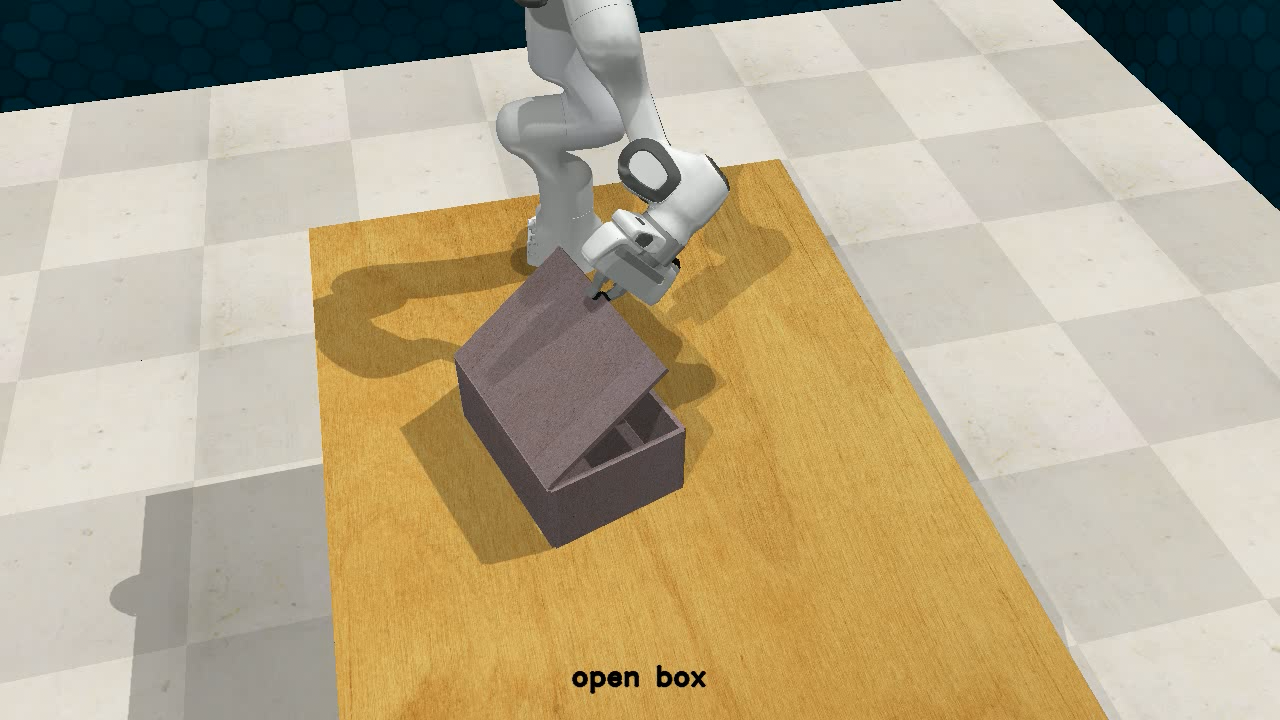} &
    \includegraphics[width=0.23\linewidth]{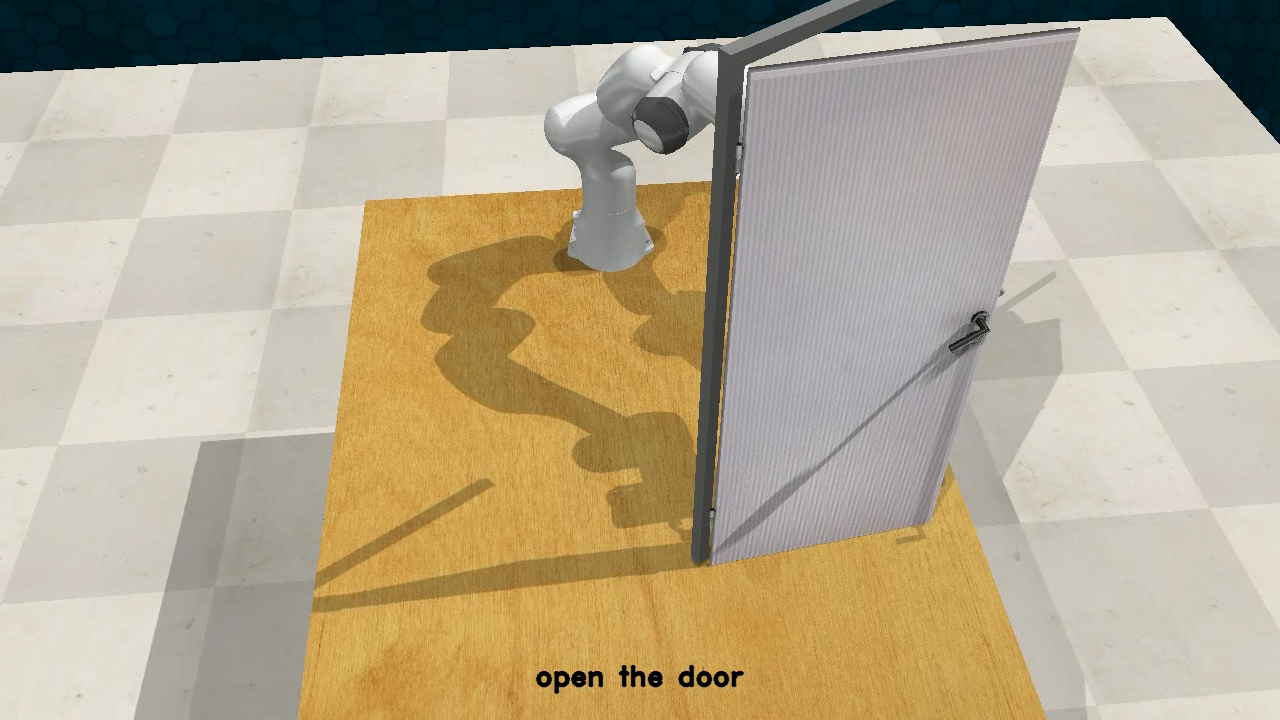} &
    \includegraphics[width=0.23\linewidth]{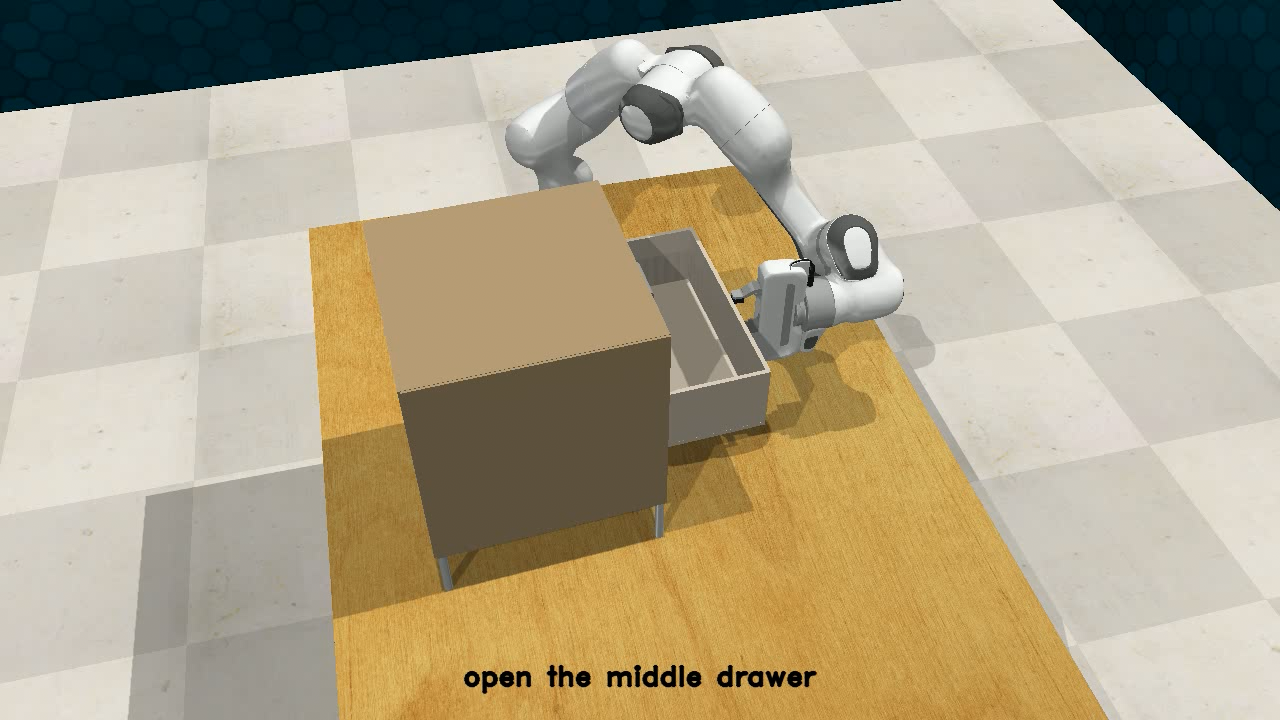} &
    \includegraphics[width=0.23\linewidth]{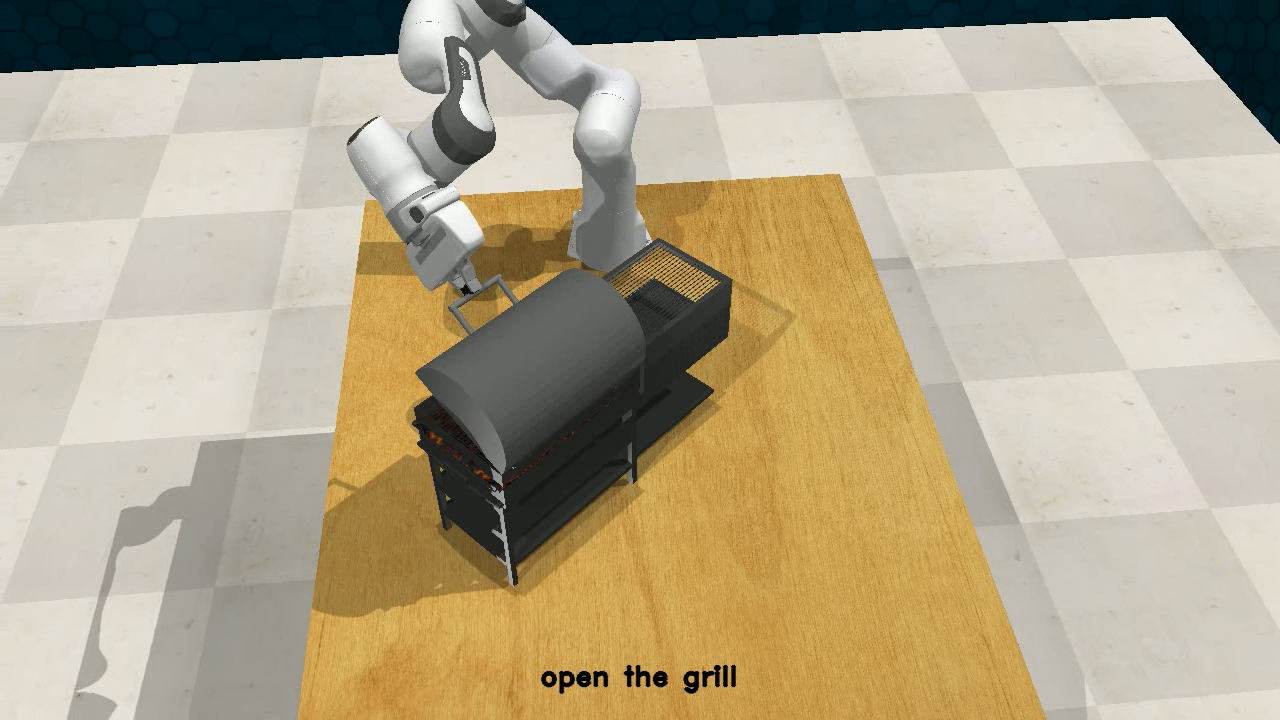} \\
    \includegraphics[width=0.23\linewidth]{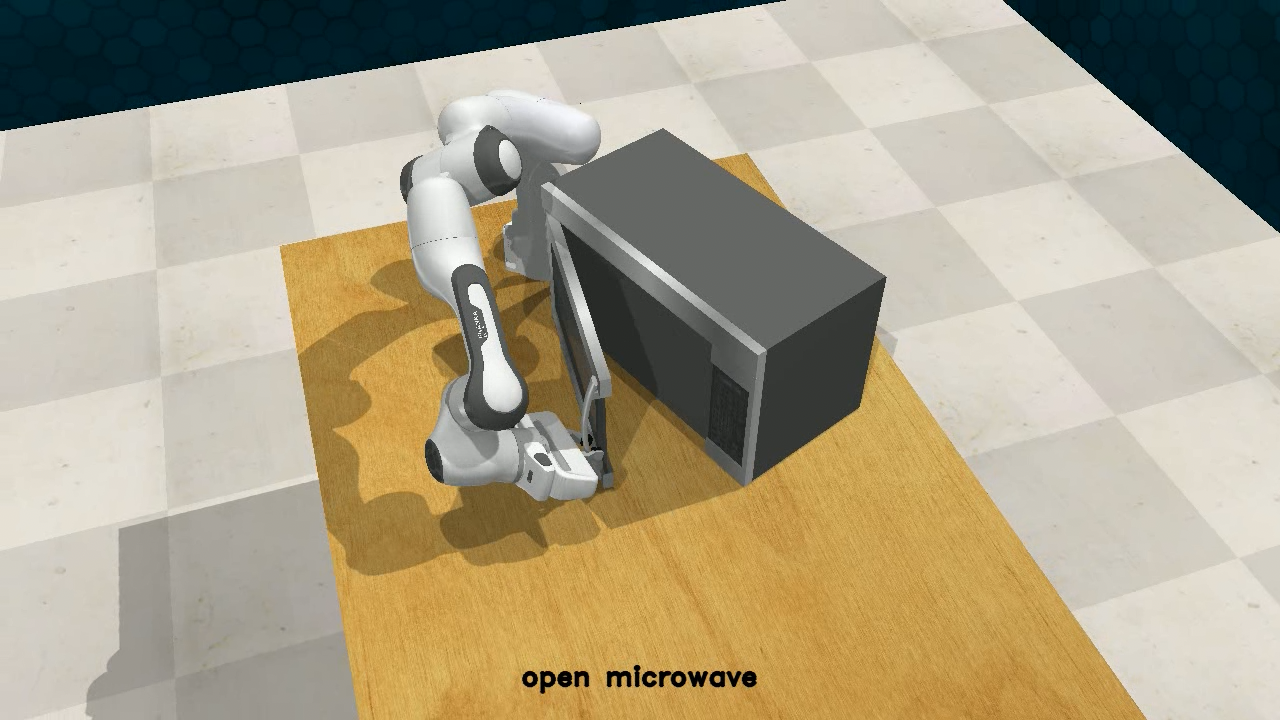} &
    \includegraphics[width=0.23\linewidth]{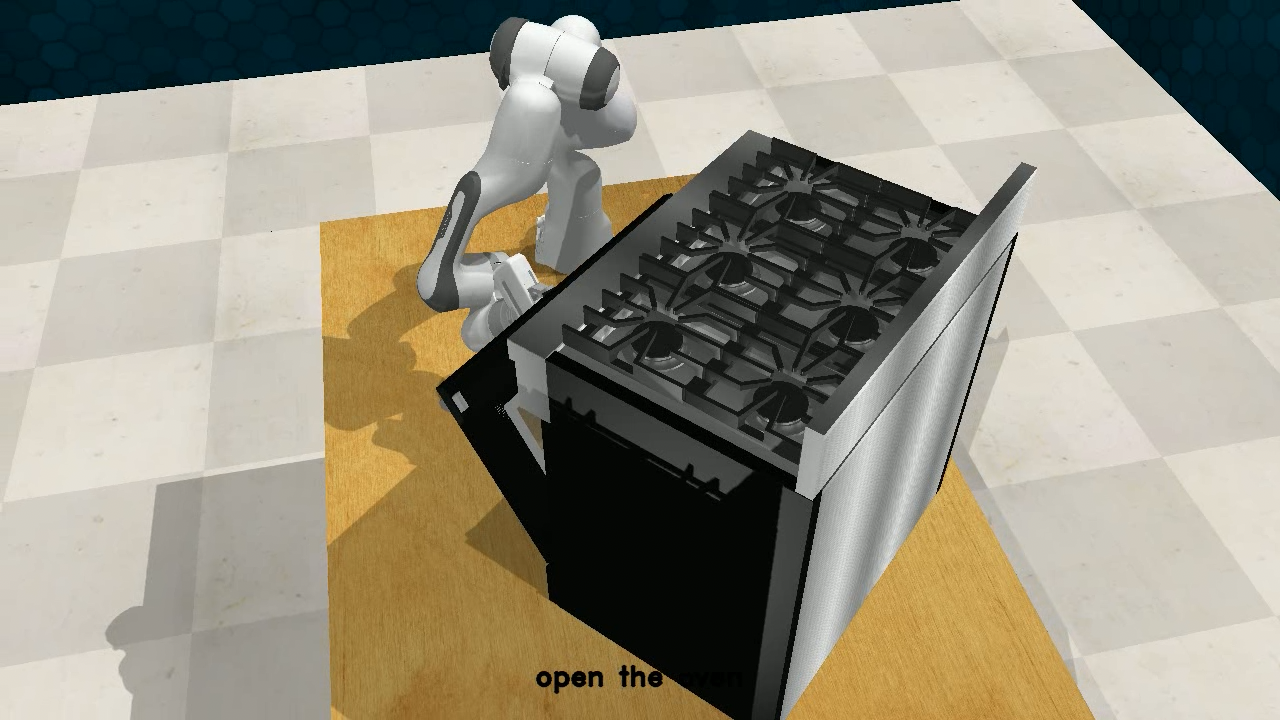} &
    \includegraphics[width=0.23\linewidth]{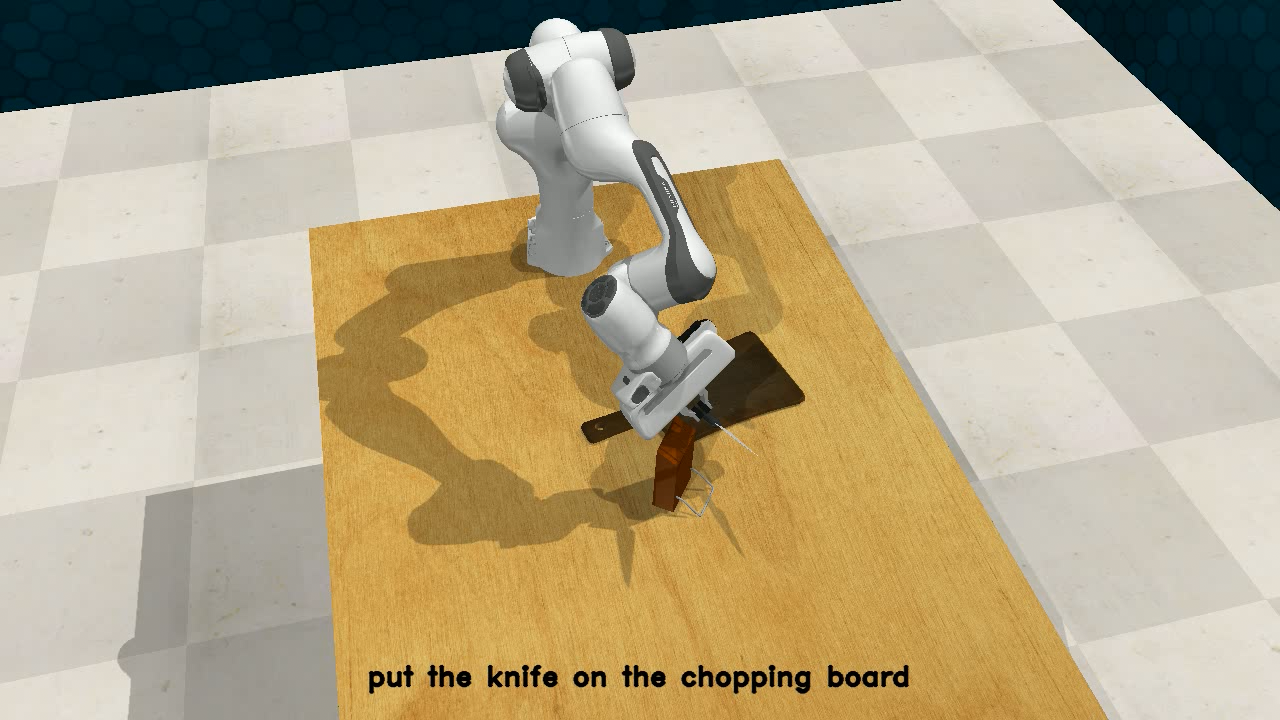} &
    \includegraphics[width=0.23\linewidth]{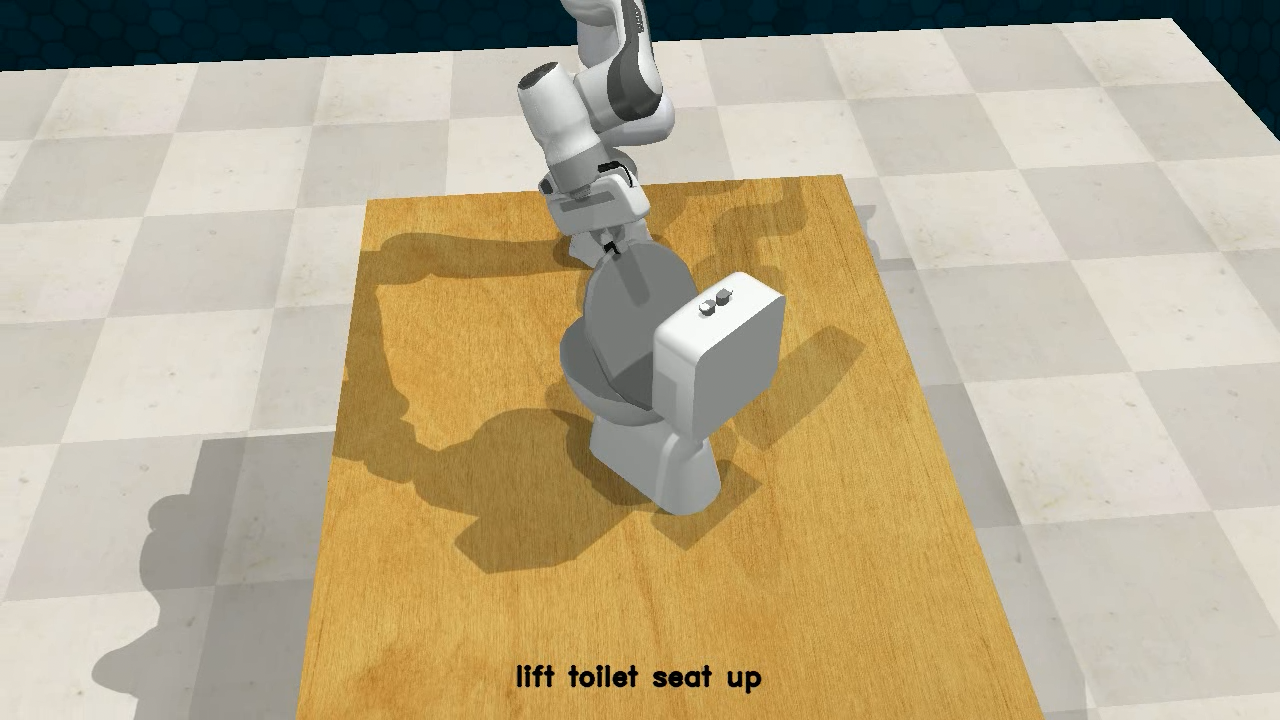}
	\end{tabular}
	\centering
	\caption{\small We focus on learning multi-task language-guided agent for robotic manipulation. Unlike a standard motion planner that only samples an arbitrary trajectory to the end pose.}
	\label{fig:illustration_example}
\end{figure*}

In this work, we introduce \fname{} (\name{}), a hierarchical multi-task agent that combines the best of both worlds. \name{} factorises a manipulation policy by chaining a high-level NBP agent with a low-level learned controller. At the high level, \name{} takes the 3D visual observations and language instructions as the inputs, and predicts a 6-DoF next-best end-effector pose. At the high level, \name{} entails the capability of understanding the visual environment and language instructions and performing long-horizon task-level decision-making. At the low level, given the high-level 6-DoF end-effector pose action as a goal, \name{} casts the control task as a context-aware 6-DoF pose-reaching task. We introduce a novel kinematics-aware low-level agent, Robot Kinematics Diffuser (\cdp{}), a diffusion-based policy~\cite{chi2023diffusionpolicy} that directly generates the motion trajectory via conditional sampling and trajectory inpainting. Specifically, instead of generating the end-effector pose trajectories as in ~\citet{chi2023diffusionpolicy,xian2023unifying} and solving the robot inverse kinematics, \cdp{} learns both end-effector pose and robot joint position diffusion, and distill the accurate but kinematics-unaware end-effector pose trajectory into the joint position trajectory via differentiable robot kinematics. \cdp{} achieves accurate trajectory generation and maximum control flexibility, while avoiding violating the robot kinematic constraints, which is a common issue of inverse kinematics solvers.

In our experiments, we empirically analyse \name{} on a wide range of challenging manipulation tasks in RLBench~\cite{james2020rlbench}. We show that \textbf{(1)} \cdp{} generally achieves a higher success rate on goal-conditioned motion generation. \textbf{(2)} The proposed hierarchical agent, \name{}, outperforms the flat baseline agents and other hierarchical variants. \textbf{(3)} \name{} can be directly trained on a real robot with only 20 demonstrations on a challenging oven opening task with a high success rate. 
\begin{figure*}[t]
    \centering
    \includegraphics[width=\linewidth]{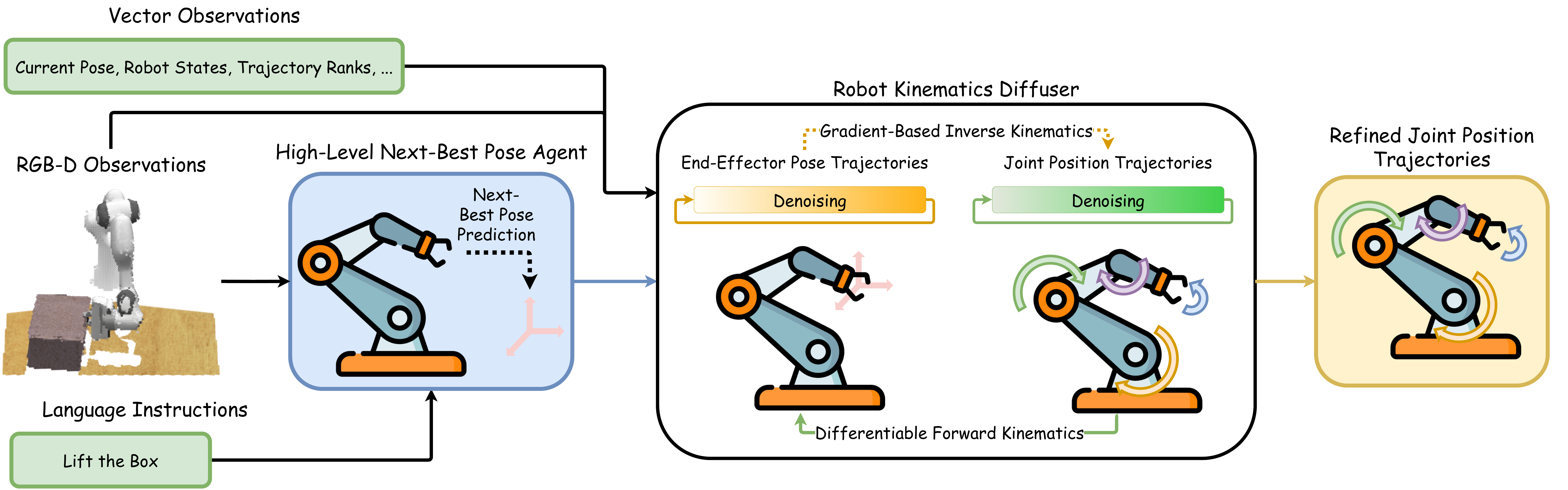}
    \caption{Overview of \fname{} (\name{}). \name{} is a multi-task hierarchical agent for kinematics-aware robotic manipulation. \name{} consists of two levels: a high-level language-guided agent and a low-level goal-conditioned diffusion policy. From left to right, the high-level agent takes in 3D environment observations and language instructions, then predicts the next-best end-effector pose. This pose guides the low-level \cdp{}. The \cdp{} subsequently generates a continuous joint-position trajectory by conditional sampling and trajectory inpainting given the next-best pose and environment observations. To generate kinematics-aware trajectories, \cdp{} distills the accurate but less flexible end-effector pose trajectories into joint position space via differentiable robot kinematics.}
    \label{fig:overview}
\end{figure*}

\section{Related Works}
\label{sec:related_works}

\subsection{End-to-End Visual Manipulation Agents}
End-to-end manipulation approaches~\cite{ james2017transferring,matas2018sim,kalashnikov2018qt,wu2019learning} make the least assumption about objects and tasks, and learn a direct mapping from RGB images to a robot action, but tend to be sample-inefficient. Two recent approaches have emerged to combat this sample inefficiency: (1) the next-best pose (NBP) action mode that learns to directly predict a distant ``keyframe"~\cite{james2022q}; (2) 3D action-value maps~\cite{james2022coarse} that aligns the 3D task space and the action space, by learning 3D voxel-based action-value maps as policies, and extracting actions by taking the coordinates of the voxel with the highest value. Such a structured action space significantly reduces the amount of data needed and the generalisation of the learned policy. In particular, built with Transformers backbones,~\citet{shridhar2023perceiver} and~\citet{gervet2023act3d} are able to take in language tokens as input, and develop language-conditioned policies. In this work, without loss of generality, we choose PerAct~\cite{shridhar2023perceiver} as our high-level language-conditioned agent for various tasks. Taking the predicted 6-DoF NBP as input, \cdp{} naturally works as a low-level policy of PerAct.
Similar to our work,~\citet{james2022lpr} combines a high-level C2F-ARM~\cite{james2022coarse} with a low-level agent that learns to rank a set of sampled trajectories by human heuristics. This approach has been shown to work on a series of challenging manipulation tasks, but it is computationally heavy and not scalable, conditioned on predefined motion generators.
We show that \name{} achieves strong multi-task manipulation capabilities with both kinematics awareness and high accuracy.

\subsection{Diffusion Models}
The diffusion model is a powerful class of generative models that learn to approximate the data distribution by iterative denoising processes. They have shown impressive results on both the conditional and unconditional image, video, and 3D object generation~\cite{ho2020denoising,rombach2022high,lu2022dpm,singer2022make,ho2022imagen,watson2022novel}. In the field of decision making, diffusion models have recently been adopted as a powerful policy class~\cite{ajay2023is,chi2023diffusionpolicy,wang2023diffusion,kang2023efficient,janner2022planning}. Specifically, Diffusion Policies~\cite{chi2023diffusionpolicy} learn to generate diverse multi-modal trajectories for robot manipulation by conditional generation with imitation learning. Concurrent with our work,~\citet{xian2023unifying} proposes ChainedDiffuser as a hierarchical agent. As we show in our experiments, the gripper-pose diffusion policy in ChainedDiffuser relies on inverse kinematics solvers to generate robot joint actions, which, however, is susceptible to prediction errors and might violate the kinematic constraints of the robot. On the contrary, the proposed CDP learns both the end-effector pose and joint position trajectories and refines the joint position trajectories by distilling the end-effector poses.

\subsection{Differentiable Physics for Decision Making}
Differentiable physics simulation constructs each simulation step as a differentiable computational graph, such that the environment steps are fully differentiable with respect to network parameters~\cite{hu2019difftaichi,huang2021plasticinelab,chen2023daxbench,xu2023efficient}. Learning decision-making policies via differentiable physics has shown to be more efficient and generalisable compared with standard non-differentiable environments~\cite{xu2022accelerated,chen2023imitation} with the physics priors as an inductive bias to the gradients. Similar to differentiable physics, we make use of the differentiable robot kinematics models~\cite{Zhong_PyTorch_Kinematics_2023} to distill the accurate but less reliable end-effector pose trajectory to the joint position space.

\section{Preliminaries}
\label{sec:preliminaries}
\subsection{Diffusion Models}

Diffusion models are a powerful family of generative models that consist of forward and backward Markov-chain diffusion processes. Consider a real data distribution $q(x)$ and a sample $x^0 \sim q(x)$ drawn from it. The forward diffusion processes adds Gaussian noise to $x^0$ in $K$ steps, which gives a sequence of noisy samples $\{x^i\}_{i=1}^K$. In DDPM~\cite{ho2020denoising}, the noise is controlled by a variance scheduler
\begin{align}
\label{eq:forward_diff}
    q(x^k|x^{k-1}) &= \mathcal{N}(x^k; \sqrt{1-\beta^k} x^{k-1}, \beta^k \vv{I}),\\
    q(x^{1:K}|x^0) &= \prod_{k=1}^{K} q(x^k|x^{k-1}).
\end{align}
where $\beta^1, \dots, \beta^K$ are scheduler parameters. Theoretically, $x^\infty$ will distribute as an isotropic Gaussian. To reconstruct distribution $q(x)$, diffusion models learn a conditional distribution $p_\theta(x^{t-1}|x^t)$ and generate new samples by
\begin{align}
\label{eq:reverse_diff}
    p_\theta(x^{0:K}) &= p(x^K) \prod_{k=1}^K p_\theta(x^{k-1}|x^k),\\
    p_\theta(x^{k-1}|x^k) &= \mathcal{N}(x^{k-1}; \vv{\mu}_\theta (x^k, k), \vv{\Sigma}_\theta(x^k, k)),\label{eqn:diff}
\end{align}
where $p(x^K) = \mathcal{N}(\vv{0}, \vv{I})$  under the condition that $\prod_{k=1}^{K}(1-\beta^{k})\approx 0$. The model can be trained by maximising the evidence lower bound (ELBO)
\begin{align}
    \mathbb{E}_{x_0}[\log p_\theta(x^0)] \ge \mathbb{E}_q\left[\log \frac{p_\theta(x^{0:K})}{q(x^{1:K}|x^0)}\right]\label{eqn:elbo}
\end{align}

In the context of decision making, diffusion policies consider a trajectory of actions $a_{1:T} = \{a(t)\}_{t=1}^T$, and learn a conditional distribution $p_\theta(a_{1:T}^{k-1}|a_{1:T}^k, \{c_i\}_{i=1}^N)$, where $\{c_i\}_{i=1}^N$ are $N$ additional conditions for policy learning, e.g., RGB observations, point cloud, robot states, etc. For simplicity, we abuse the notation and denote $a_{1:T}^k$ as $\vv{a}^k$.

\subsection{Differentiable Kinematics}
Differentiable simulation aims to encode the physics simulation steps as a differentiable computational graph. Take a simple point-mass system as an example.
\begin{align}
    y_t = y_{t-1} + \Delta t * v_t,\quad v_{t+1} = v_t + \Delta t * \frac{F}{m}
\end{align}
where the force $F$ is the input to the system, $m$ is the mass, $v$ is the speed, and $y$ is the position of the point. Importantly, such a system is differentiable and we can optimise the input force $f$ by the gradients from positions $y$. Similarly in the context of robotics, conditioned on a predefined URDF model of a robot, the end-effector pose $\vv{s}_p$ of a robot can be obtained by a differentiable forward kinematics function $f_{K}$ as $\vv{s}_p = f_K(\vv{s}_j)$, where  $\vv{s}_j$ is the joint angles. Thus, given a loss function $L(\vv{s}_p)$ over the gripper poses, the joint positions can be directly updated by gradients $\frac{\partial L(\vv{s}_p)}{\partial \vv{s}_j}$.

\section{\fname{}}

The overall pipeline of \name{} is shown in Fig.~\ref{fig:overview}. 

\textbf{Problem Definition}. We aim to learn a \name{} policy $\pi(\vv{a}\mid o, l)$, which processes the RGB-D observation $o$ and language instruction $l$, specifying the task, to predict a hybrid action $\vv{a}$. Here, $\vv{a}$ consists of a trajectory $\vv{a}_\joint = \{a(0), a(1), \dots, a(T)\}$ and gripper opening / closing action $a_\mathrm{grip}$, where $T$ is the trajectory length and $a(i)\in \mathbb{R}^N$, with $N$ denoting the number of the robot joints. For brevity, we symbolise actions $\vv{a}$ without temporal index in the episode. 

\textbf{Factorised Hierarchical Policy.} To tackle long-horizon context-aware manipulation tasks, we factorise the policy $\pi(\vv{a}\mid o, l)$ into a hierarchical structure. Specifically, $\pi(\vv{a} \mid o, l) = \pi_\mathrm{high}(a_\mathrm{high}\mid o, l)\circ \pi_\mathrm{low}(\vv{a}\mid o, a_\mathrm{high})$. Here, the high-level action $a_\mathrm{high} = (a_\pose, a_\mathrm{grip}) \sim \pi_\mathrm{high}$, consists of (1) the end-effector pose action $a_\pose = (a_\mathrm{trans}, a_\mathrm{rot})$, with translation action $a_\mathrm{trans}\in \mathbb{R}^3$ and quaternion rotation action $a_\mathrm{rot}\in \mathbb{R}^4$; and (2) a binary gripper action $a_\mathrm{grip}\in \mathbb{R}$. Conditioned on the high-level action $a_\mathrm{high}$, we parameterise the low-level policy $\pi_\mathrm{low}(\vv{a}\mid a_\mathrm{high}, o)$ with \cdp{}, and learn to generate accurate joint position trajectories $\vv{a}_\mathrm{low} = \vv{a}_\joint$. Such a factorisation offloads the complex and expensive task-level understanding from language instructions to the high-level agent, leaving only control to be learned by a simple, goal-conditioned low-level agent. During inference, \name{} works in a sequential manner and we take $\vv{a} = \{\vv{a}_\joint, a_\mathrm{grip}\}$ as the output.

\subsection{Dataset Preparation}
We assume access to a multi-task dataset $\mathcal{D} = \{\xi_i\}_{i=1}^{N_D}$, containing a total of $N_D$ expert demonstrations paired with $\mathcal{D}_l = \{l_i\}_{i=1}^{N_D}$ language descriptions. Note that a single task might have multiple variations, each with different description, e.g., ``open the middle drawer" or ``open the bottom drawer". Each demonstration $\xi = \{\vv{a}_\mathrm{demo}, \vv{o}_\mathrm{demo}\}$, consists of an expert trajectory $\vv{a}_\mathrm{demo}$ and resulting observation $\vv{o}_\mathrm{demo}$. To enable the training of both the high-level policy $\pi_\mathrm{high}$ and the low-level \cdp{} $\pi_\mathrm{low}$, the action $\vv{a}_\mathrm{demo}$ includes: (1) end-effector poses $\vv{a}_\pose$; (2) gripper opening / closing  action $\vv{a}_\mathrm{grip}$; and (3) joint positions $\vv{a}_\joint$. The observation $\vv{o}_\mathrm{demo}$ includes multi-view calibrated RGB-D camera observations and robot states. 

\textbf{Keyframe Discovery.} Referring to prior works~\cite{james2022q,james2022coarse}, training the high-level agent on all trajectory points is inefficient and instead we apply a Keyframe discovery method introduced in~\citet{james2022q}. Scanning through each trajectory $\xi$, we extract a set of $K_\xi$ keyframe indices $\{k_i\}_{i=1}^{K_\xi}$ that capture the principal bottleneck end-effector poses. Specifically, a frame is considered as a keyframe if (1) the joint velocity is close to 0; and (2) the gripper open / close state remains unchanged. Unlike prior works, which only keep keyframes for training, we maintain the keyframe indices and extract different segments of data to train both high-level and low-level agents. The details will be discussed in the following sections.

\subsection{High-Level Next-Best Pose Agent}
For the high-level policy $a_\mathrm{high} = (a_\pose, a_\mathrm{grip})\sim \pi_\mathrm{high}(a\mid o, l)$, we utilise a next-best pose agent~\cite{james2022q} with structured action representations. In this work, to parameterise $\pi_\mathrm{high}$ and fulfil this objective, we employ Perceiver-Actor (PerAct)~\cite{shridhar2023perceiver}. PerAct is a language conditioned Behaviour Cloning (BC) agent with Transformer~\cite{vaswani2017attention} backbones. PerAct achieves its high sample-efficiency, generalisability and accuracy through the use of a high-resolution voxel scene representations to predict 3D voxel-based action-value maps. To tackle the large number of visual and language tokens, PerAct adopts PerceiverIO~\cite{jaegle2021perceiver}, which encodes the inputs with a small set of latent vectors and reduces the computational complexity. 

\textbf{Action Spaces}. PerAct uses discrete action spaces for all action heads, including (1) a discrete policy head over the voxels for $a_\mathrm{trans}$ and (2) a pair of discrete policies for $a_\mathrm{rot}$ and $a_\mathrm{grip}$. Continuous actions are reconstructed by converting the discrete indices according to the action space ranges. 

\textbf{Model Training.} For the high-level agent, we use only the keyframes for training. In addition, following~\citet{shridhar2023perceiver}, we use demo augmentation and translation augmentation to generate more samples. The network is optimised by behaviour cloning losses, i.e., cross-entropy losses in the discrete action space
\begin{align}
    \mathcal{L}_\mathrm{high} = - \mathbb{E}_{k\sim \xi, \xi\sim\mathcal{D}}\left[ \log \pi_\mathrm{high}(a_\mathrm{demo}(k) \mid o, l) \right]
\end{align}
where $a_\mathrm{demo}(k)$ is the expert action of the keyframe $k$.

\subsection{Low-Level \cdp{}}\label{sect:diffuser}
Given the predicted high-level action $a_\mathrm{high}$, we perform conditional trajectory generation with \cdp{} through denoising diffusion processes.
Standard diffusion policy for robotic manipulation considers end-effector pose diffusion
\begin{align}
&p_\theta (\vv{a}_\pose^{k-1} \mid \vv{a}_\pose^k, C_\pose)\nonumber\\
&= \mathcal{N}(\vv{a}_\pose^{k-1}; \vv{\mu}_\theta (\vv{a}_\pose^k, C_\pose, k), \vv{\Sigma}_\theta(\vv{a}_\pose^k, C_\pose, k))
\end{align}
where $C_\pose$ consists of the conditional variables, including the known start pose $a(0)_\pose^0$, the predicted next-best pose $\hat{a}_\pose^0(T)$ by the high-level agent, the low-dimensional state $s$ of the robot, the end-effector pose, the gripper open amount, and the point cloud of the environment $\vv{v}$. 

Besides using the start and next-best pose as conditional variables to the networks, we inpaint the trajectory with the start pose and the predicted next-best pose at each denoising step. This end-effector pose diffusion allows the inpainting operation to act as a hard constraint for the diffusion process, which guarantees the last step in the trajectory will always be aligned with the output of the high-level agent. 

Prior to execution, the end-effector pose trajectory must undergo processing by an inverse kinematics (IK) solver to determine corresponding joint positions. However, the predicted end-effector pose trajectory lacks kinematics awareness and there is a high likelihood of it violating the kinematic constraints. Consider, for example, that each step of the predicted trajectory has a probability $p$ to violate the IK constraints. For a trajectory of length $T$, the probability of the trajectory might violate the constraint is $p_\mathrm{error} = 1 - (1 - p) ^T$, and $\lim\limits_{T\rightarrow \infty} p_\mathrm{error} = 1$. As we show in our experiments, IK error contributes to most of the failure cases in end-effector pose trajectory diffusion.

\textbf{Kinematics-Aware Diffusion}. As an alternative to using IK solvers, the robot could be operated through joint position control. This approach provides direct and complete control of the robot. However, learning a trajectory diffusion model in the joint position space is challenging. In the case of end-effector pose diffusion models, we can impose accurate and strong constraints with the predicted next-best pose $\hat{a}_\pose^0(T)$. However, for an over-actuated 7-DoF robot arm, a 6-DoF end-effector pose $\hat{a}_\pose^0(T)$ might have an infinite number of corresponding joint positions $\hat{a}_\joint^0(T)$, which makes it difficult to perform inpainting for joint position diffusion. As we show in our experiments, the naive joint position diffusion model tends to be less accurate for goal-conditioned control, especially for the end poses.

To tackle this issue, we introduce Robot Kinematics Diffuser (\cdp{}). Similar to~\citet{xian2023unifying}, \cdp{} learns an end-effector pose diffusion model $p_\theta (\vv{a}_\pose^{k-1}\mid \vv{a}_\pose^k, C_\pose)$ which generates accurate but less reliable end-effector pose trajectories. \cdp{} further learns an additional joint position diffusion model
\begin{align}
&p_\phi (\vv{a}_\joint^{k-1} \mid \vv{a}_\joint^k, C_\pose)\nonumber\\
&= \mathcal{N}(\vv{a}_\joint^{k-1}; \vv{\mu}_\theta (\vv{a}_\joint^k, C_\pose, k), \vv{\Sigma}_\theta(\vv{a}_\joint^k, C_\pose, k))
\end{align}
where we use the same set of conditional variables $C_\pose$ for conditional generation, but for inpainting, we only fix the initial joint action $a_\joint^0(0)$.

For action trajectories sampled from each learned policy, $\vv{a}_\pose^0 \sim p_\theta(\vv{a}_\pose^0\mid \vv{a}_\pose^1, C_\pose)$ and $\vv{a}_\joint^0\sim p_\phi(\vv{a}_\joint^0\mid \vv{a}_\joint^1, C_\pose)$, we can build such a mapping by treating the differentiable robot kinematics model $f_K$ as a function $\hat{\vv{a}}_\pose^0  = f_K(\vv{a}_\joint^0)$. During inference, initialised with a near-optimal solution $\vv{a}_\joint^0$, we can optimise the joint positions $\vv{a}_\joint^0$ to predict end-effector poses $\hat{\vv{a}}_\pose^0$ that are close to $\vv{a}_\pose^0$ using gradients
\begin{align}
    \vv{a}_\joint^0 \leftarrow \vv{a}_\joint^0 - \alpha \frac{\partial \parallel \vv{a}_\pose^0 - \hat{\vv{a}}_\pose^0 \parallel}{\partial \vv{a}_\joint^0},
\end{align}
where $\alpha$ is the learning rate. This gives a trajectory $\vv{a}_\joint^{0*}$ that does not violate the kinematics constraint of the robot while achieving a high accuracy for manipulation tasks.

\textbf{Networks.} The low-level \cdp{} takes as input the start pose, the end pose, the RGB-D image of the first step observation, a vector of the robot low-dimensional states, and the trajectory rank. For the RGB-D image, we first convert it to a point cloud in the world frame and extract the features with PointNet++~\cite{qi2017pointnet++}; for the other vector features, we use 4 layers of MLP. For the temporal encoding network, we found a temporal Conv1D UNet used in~\citet{janner2022planning} performs well and has no clear performance gap between the commonly adopted Transformer backbones. 

\textbf{Model Training.} When training diffusion models, we aim to maximise the ELBO of the dataset (Eqn.~\ref{eqn:elbo}). However, taking the predicted next-best poses from the high-level policy $\pi_\mathrm{high}$ is inefficient as the prediction might be sub-optimal and slow. To alleviate this issue, for each demonstration $\xi$, we construct sub-trajectories $\{\xi(i)\}_{i=1}^K$ by chunking the trajectory $\xi$ with the detected keyframe indices $\{k_i\}_{i=1}^{K_\xi}$. Next, we relabel each keyframe as a sub-goal of the training trajectory. This aligns with the training of the high-level agent $\pi_\mathrm{high}$, and in practice, $\pi_\mathrm{high}$ and $\pi_\mathrm{low}$ can be optimised at the same time. The relabeling idea also resembles the Hindsight Experience Replay~\cite{andrychowicz2017hindsight}, which has been shown to be effective in learning hierarchical policy learning~\cite{levy2017learning,nguyen2022hierarchical}. Specifically, we have

\begin{align}
    \mathcal{L}_\mathrm{low} &= -\beta_1\mathcal{L}_\pose - \beta_2 \mathcal{L}_\joint - \beta_3 \mathcal{L}_{\joint \rightarrow \pose}\nonumber\\
    \mathcal{L}_\pose &= \mathbb{E}_{q, \xi(i)\sim \mathcal{D}}\left[ \log \frac{p_\theta(\vv{a}_\pose^{0:K}\mid \xi(i))}{q(\vv{a}_\pose^{1:K}\mid \vv{a}_\pose^0, \xi(i))}\right]\nonumber\\
    \mathcal{L}_\joint &= \mathbb{E}_{q, \xi(i)\sim \mathcal{D}}\left[ \log \frac{p_\phi(\vv{a}_\joint^{0:K}\mid \xi(i))}{q(\vv{a}_\joint^{1:K}\mid \vv{a}_\joint^0, \xi(i))}\right]\nonumber\\
    \mathcal{L}_{\joint\rightarrow\pose} &= \mathbb{E}_{q, \xi(i)\sim \mathcal{D}}\left[ \log \frac{p_\phi(\vv{a}_\pose^{0:K}\mid \xi(i))}{q(\vv{a}_\pose^{1:K}\mid \vv{a}_\pose^0, \xi(i))}\right],
\end{align}
where $\beta_1$, $\beta_2$, and $\beta_3$ are weighting parameters and $\xi(i)$ is a sub-trajectory sampled from the dataset with start and end relabeled to two nearby keyframes. In particular, $\mathcal{L}_{\joint\rightarrow\pose}$ is made possible by predicting the end-effector poses from joint positions via differentiable kinematics $\hat{\vv{a}}_\pose^{0:K} = f_K(\vv{a}_\joint^{0:K})$. This allows us to train a joint-position trajectory which better regularizes the joint positions with the kinematics as an inductive bias.

\textbf{Trajectory Ranking}. During training, most of the manipulation algorithms use sampling-based motion planners whose trajectories might be sub-optimal. 
In \cdp{}, we propose to add an additional conditional variable for each sub-trajectory, a trajectory rank $r_\xi = \frac{d_\mathrm{Euclidean}}{d_\mathrm{travel}}$, where $d_\mathrm{Euclidean}$ is the Euclidean distance between the start and end pose and $d_\mathrm{travel}$ is the travelled distance between the start and end pose. Intuitively, an optimal trajectory, ignoring the kinematics constraint of the robot, should have $r_\xi = 1$. To encourage \cdp{} to generate near-optimal trajectories, we set $r_\xi = 1$ during inference. An analysis of the influence of trajectory ranking is in the appendix.

\subsection{Practical Implementation Choices}
For the high-level agent $\pi_\mathrm{high}$, different from the past work~\cite{james2022lpr, shridhar2023perceiver}, we ignore the $a_\mathrm{collsion}$, which is a binary variable used to indicate whether the motion planner should perform collision avoidance because the low-level \cdp{} is trained to generate collision-aware optimal trajectories. For the low-level agent, different from most of the diffusion models that learn to predict a noise prediction model and learn to reconstruct the noise during the denoising steps, we follow~\citet{ramesh2022hierarchical} and observe that empirically directly predicting the original actions $\vv{a}^0_\pose$ and $\vv{a}^0_\joint$ is giving better performance. Besides, when truncated by the keyframe indices, the sub-trajectories might have different lengths. To tackle this issue, we resample each trajectory into a length of 64 for batched training. More implementations and discussions are in the appendix.

\section{Experiments}\label{sec:experiments}

\begin{table*}[t]
\caption{Success Rates (\%) on RLBench Tasks. For \textcolor{red}{red} tasks, we expect no improvement of HDP over baselines; with \textcolor{blue}{blue} tasks, we expect HDP to outperform many of the baselines.}
\label{tab:main_result}
\small
\centering
 \fontsize{7}{8}\selectfont
\begin{tabular}{@{}ccccccccccccc@{}}
\toprule
                  & \textcolor{red}{\thead{reach\\ target}} & 
                  \textcolor{red}{\thead{take lid off\\ saucepan}} & 
                  \textcolor{red}{\thead{pick\\ up cup}} & 
                  \textcolor{blue}{\thead{toilet\\seat up}} & 
                  \textcolor{blue}{\thead{open\\box}} & 
                  \textcolor{blue}{\thead{open\\door}} & 
                  \textcolor{blue}{\thead{open\\drawer}} & 
                  \textcolor{blue}{\thead{open\\grill}} & 
                  \textcolor{blue}{\thead{open\\microwave}} & 
                  \textcolor{blue}{\thead{open\\oven}} & 
                  \textcolor{blue}{\thead{knife on\\board}} & overall \\ 
\cmidrule(lr){1-1}\cmidrule(lr){2-13}
ACT               &   50           &            45           &        46     &        6        &     12     &      5     &       26      &     1       &       11         &      0     &         0       &    18.36     \\
Diffusion Policy  &       43       &          25             &      24       &      5          &     4     &     22      &      28       &    9        &       7         &    0       &        0        &    15.18     \\
PerAct + Planner  & 100          & \textbf{100}                   & \textbf{86}          & 0              & 0        & 64        & 68          & 54         & \textbf{32}             & 0         & \textbf{76}             & 57.72   \\
PerAct + Planner + Bezier      & 96           & 100                   & 72          & 80             & 8       & 48          & 84             & 76            & 20               & 4          & 36               & 56.73        \\
PerAct + Diffuser & 100          & 94                    & 84          & 80             & 82       & 88        & 84          & 82         & 20             & 18        & 52             & 71.27   \\
\cmidrule(lr){1-1}\cmidrule(lr){2-13}
HDP               & \textbf{100}          & 96                    & 82          & \textbf{86}             & \textbf{90}       & \textbf{94}        & \textbf{90}          & \textbf{88}         & 26             & \textbf{58}        & 72             & \textbf{80.18}   \\ \bottomrule
\end{tabular}
\end{table*}

In our experiments we show the following: \textbf{(1)} \name{} outperforms the state-of-the-art methods across all RLBench tasks; \textbf{(2)} in general, hierarchical agents outperform simple low-level continuous control policies; and \textbf{(3)} task-aware planning is important for many manipulation tasks, in particular those involving articulated objects.

In addition to this, we perform a series of ablation studies and show: \textbf{(1)} IK errors contribute to the majority of the failure cases of end-effector pose diffusion policy; \textbf{(2)} joint position diffusion is less accurate without the access to last joint position inpainting; and \textbf{(3)} 3D information and the corresponding feature extraction module are critical to the performance of \cdp{}.

Finally, we show \name{} is capable of solving challenging real-world tasks efficiently and effectively on an open oven task with only 20 demonstrations.

For all simulation experiments, we use 100 demonstrations from RLBench~\cite{james2020rlbench} for each task and train for 100K iterations.
On a real robot, we show \name{} can learn efficiently and effectively with only 20 demonstrations.

\begin{figure}[t]
	\centering 
	\begin{tabular}{ccc}
    \includegraphics[width=0.3\linewidth]{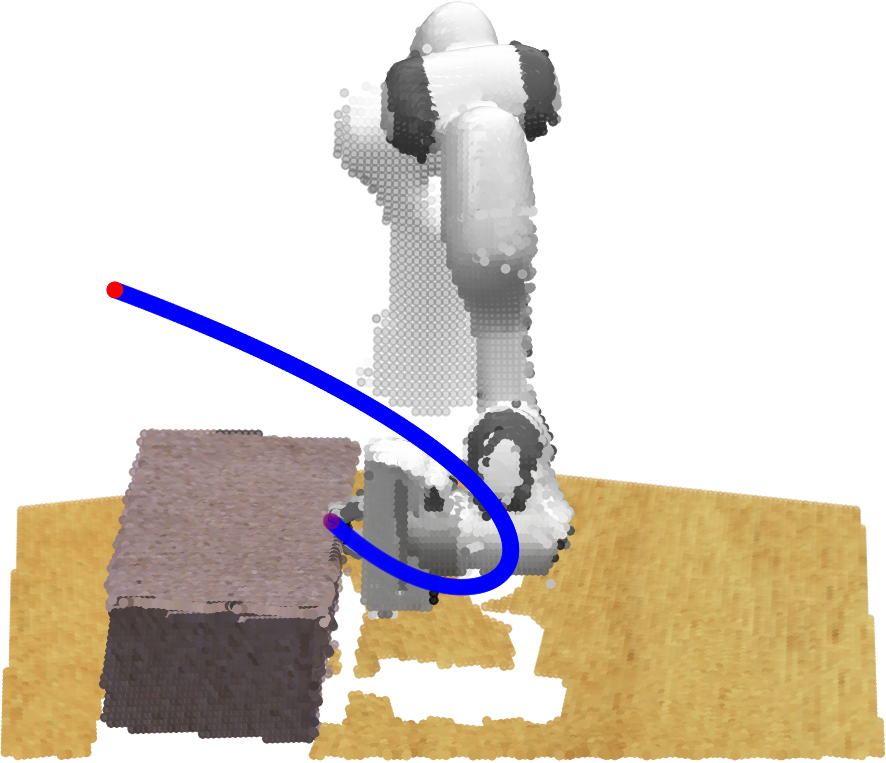} &
    \includegraphics[width=0.3\linewidth]{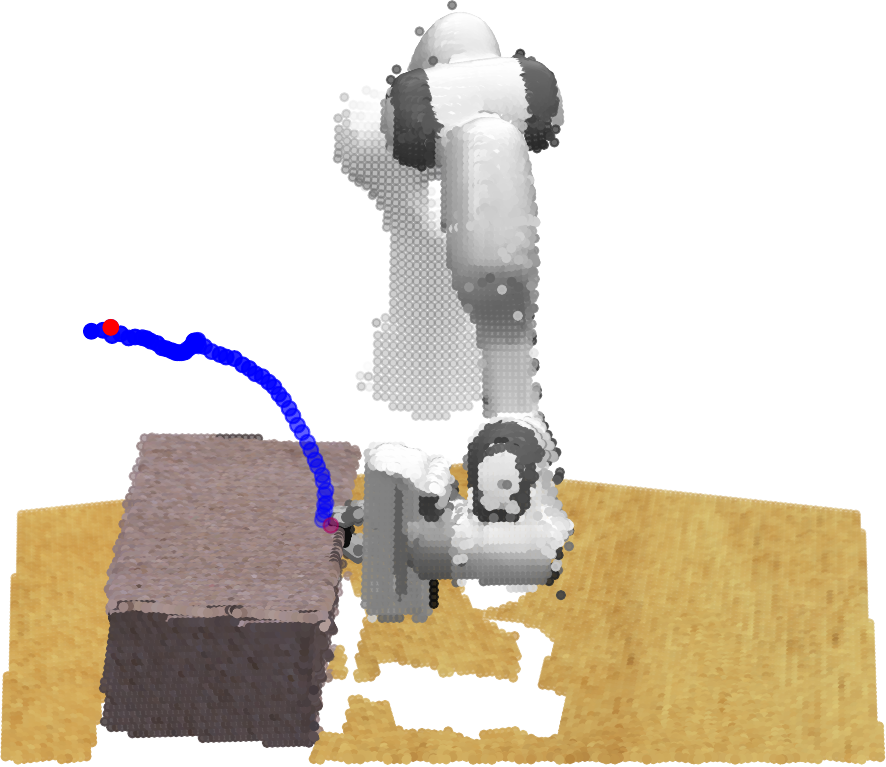} &
    \includegraphics[width=0.3\linewidth]{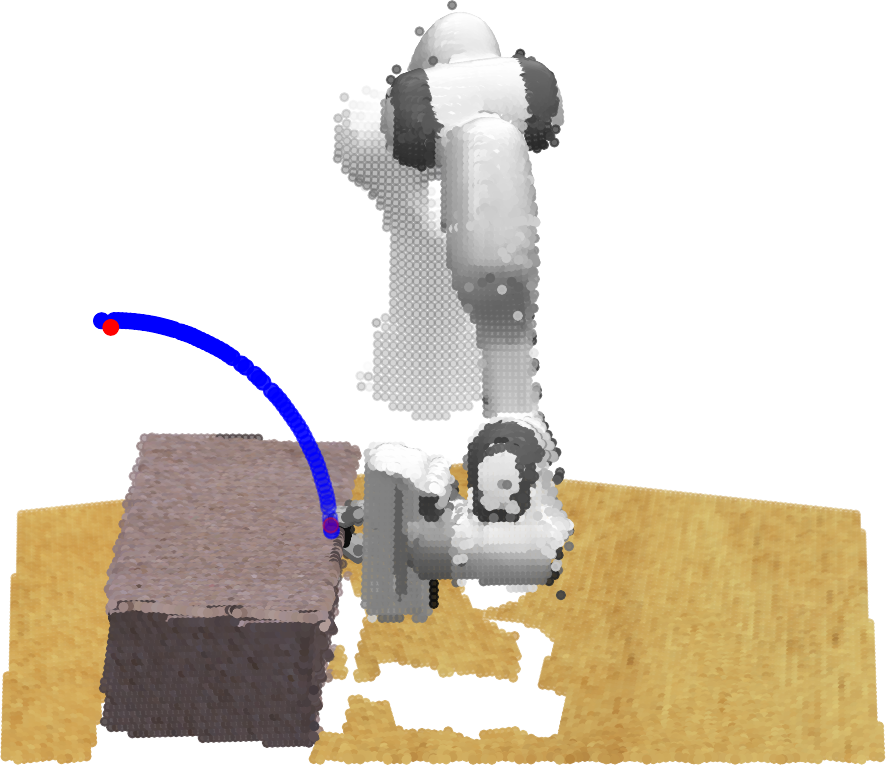} \\
    \small (a) RRT & \small (b) Joint Position & (c) \small \cdp{}
\end{tabular}
	\centering
	\caption{\small Trajectory visualisations of the open box task. 
 }\label{fig:traj_vis}
\end{figure}
\subsection{Trajectory Visualisations}\label{sect:vis}
Firstly, we aim to understand why learning a low-level controller is necessary. In Fig.~\ref{fig:traj_vis}, we visualise the trajectory of an open box task in RLBench. RRT learns a trajectory that correctly reaches the goal pose. Nevertheless, without understanding the task context, the trajectory generated by RRT will cause the lid of the box to fall from the gripper. To visualise the joint position trajectories of both the vanilla joint position diffusion policy and \cdp{}, we further predict the end-effector poses from the joint positions. Although the joint position diffusion policy understands the task context, without direct inpainting with the next-best joint position, the trajectory will be less accurate. \cdp{} distills the accurate end-effector poses to the joint positions via differentiable kinematics, which achieves both high prediction accuracy and kinematic awareness.

\subsection{Simulation Experiments}
We aim to compare \name{} against (1) the state-of-the-art low-level control behaviour cloning agents, including \textbf{ACT}~\cite{zhao2023learning} and the vanilla \textbf{Diffusion Policy}~\cite{chi2023diffusionpolicy}; (2) the high-level next-best-pose agent with a fixed local planner, PerAct. In addition, we aim to demonstrate the benefit of the proposed \cdp{} against alternatives, including: (1) \textbf{Planner}: a hybrid planner of fixed linear paths and standard RRT, which is the default setup used in RLBench; (2) \textbf{Planner + Bezier}: in which an additional head is added to the PerAct backbone with a discrete output trained to choose the most appropriate trajectory generation method at each episode step, akin to Learned Path Ranking (LPR)~\cite{james2022lpr} in the behaviour cloning setting; (3) \textbf{Diffuser}: the vanilla Diffuser~\cite{janner2022planning} framed as a goal-conditioned joint-position diffusion model. More details of the baseline algorithms are available in the appendix.
We choose 11 RLBench tasks ranging from simple context-unaware grasping tasks to challenging tasks that require interacting with articulated objects. We present the results in Tab.~\ref{tab:main_result} and make the following observations.

\textit{\name{} outperforms the state-of-the-art methods across RLBench tasks.} As shown in Tab.~\ref{tab:main_result}, \name{} achieves an overall 80.2\% success rate across 11 RLBench tasks. In particular, we observe on simple tasks (\textcolor{red}{red}), that require no accurate trajectory control, most of the baselines have achieved a competent performance. However, when it comes to the more challenging tasks (\textcolor{blue}{blue}), \name{} maintains its performance while the baselines mostly fail, due to either a lack of understanding in the task context or inaccurate motion trajectory generation.

\textit{Hierarchical agents outperform simple low-level continuous control policies.} Comparing ACT and the vanilla Diffusion Policy with hierarchical agents, we observe that hierarchical agents consistently outperform the former.  Empirically, both ACT and the Diffusion Policy fail to accurately detect intermediate keyframes, such as the handle of a drawer or an oven. This error is amplified due to distribution shift, which is a common issue of behaviour cloning agents in long-horizon tasks. In contrast, the hierarchical agent, with PerAct at the high level, achieves better generalisation and simplifies the optimisation task of the low-level agent. When trained on a multi-task setting, both ACT and Diffusion Policy fail to manage different skills and generalise to unseen test examples. However, all algorithms achieve a low performance on the open microwave task. We observe that this task has a highly diverse final end-effector pose distribution, which causes the high-level policy to have a high variance and generate inaccurate next-best poses. This error is then propagated to the low-level agents. Further exploration of this issue is left for future study.

\textit{Learned low-level agents achieve better performance than motion planners.} In particular, we note that even with accurate predictions of the next-best pose, a lack of task understanding by the planner often leads to trajectories deviating from the desired optimal trajectory. For instance, while PerAct + Planner achieves 0\% success rate on the open box task it regularly succeeds in grasping the box lid. The predicted trajectory consistently exceeds the turning radius of the lid hinge, leading to the failure. This issue is exacerbated by strict kinematic limitations. For example, in the same task, PerAct + Planner + Bezier performs poorly because, unlike in the lift toilet seat task, the smooth opening curves, prompted by the additional head of PerAct, are kinematically infeasible. On the contrary, the learned trajectories capture the task context as demonstrated by the data and result in superior performance on a greater number of tasks.

\begin{table*}[!htb]
\caption{Ablation Study: Success Rates (\%) / IK Error Rates (\%) of low-level agents with the ground-truth next-best poses. For \textcolor{red}{red} tasks, we expect no improvement of HDP over baselines; with \textcolor{blue}{blue} tasks, we expect HDP to outperform many of the baselines.}
\label{tab:ablation}
\small
 \fontsize{7}{8}\selectfont

\centering
\begin{tabular}{ccccccccccccc}
\toprule
                  & \textcolor{red}{\thead{reach\\ target}} & 
                  \textcolor{red}{\thead{take lid off\\ saucepan}} & 
                  \textcolor{red}{\thead{pick\\ up cup}} & 
                  \textcolor{blue}{\thead{toilet\\seat up}} & 
                  \textcolor{blue}{\thead{open\\box}} & 
                  \textcolor{blue}{\thead{open\\door}} & 
                  \textcolor{blue}{\thead{open\\drawer}} & 
                  \textcolor{blue}{\thead{open\\grill}} & 
                  \textcolor{blue}{\thead{open\\microwave}} & 
                  \textcolor{blue}{\thead{open\\oven}} & 
                  \textcolor{blue}{\thead{knife on\\board}} & overall \\ 
\cmidrule(lr){1-1}\cmidrule(lr){2-13}
RRT             & 100 / 0 & 100 / 0 & 95 / 0 & 0 / 0   & 0 / 0   & 0 / 0  & 0 / 0   & 0 / 0  & 0 / 0  & 0 / 0  & 0 / 0  & 26.82 / 0 \\
Pose Diffusion  & 100 / 0 & 85 / 6  & 93 / 0 & 93 / 4  & 88 / 8  & 24 / 68 & 3 / 88   & 64 / 22 & \textbf{98 / 0} & 9 / 62  & 82 / 12 & 67.18 / 24.55 \\
Joint Diffusion & 100 / 0 & 100 / 0 & 91 / 0 & 95 / 0  & 100 / 0 & 74 / 0 & 15 / 0  & 62 / 0 & 75 / 0 & 13 / 0 & 85 / 0 & 73.64 / 0 \\
RKD-RGB         & 100 / 0 & 96 / 0  & 78 / 0 & 40 / 0  & 98 / 0  & 94 / 0 & 78 / 0  & 36 / 0 & 80 / 0 & 0 / 0  & 94 / 0 & 72.18 / 0 \\
RKD-ResNet      & 100 / 0 & 100 / 0 & 95 / 0 & 92 / 0  & 100 / 0 & 93 / 0 & 100 / 0 & 86 / 0 & 21 / 0 & 43 / 0 & 88 / 0 & 83.45 / 0 \\
\cmidrule(lr){1-1}\cmidrule(lr){2-13}
\cdp{}     & \textbf{100 / 0} & \textbf{100 / 0} & \textbf{98 / 0} & \textbf{100 / 0} & \textbf{100 / 0} & \textbf{95 / 0} & \textbf{100 / 0} & \textbf{90 / 0} & 88 / 0 & \textbf{75 / 0} & \textbf{94 / 0} & \textbf{94.55 / 0}  \\
\bottomrule
\end{tabular}
\end{table*}
\begin{figure*}[!htb]
	\centering
	\begin{tabular}{ccccc}
    \includegraphics[width=0.15\linewidth]{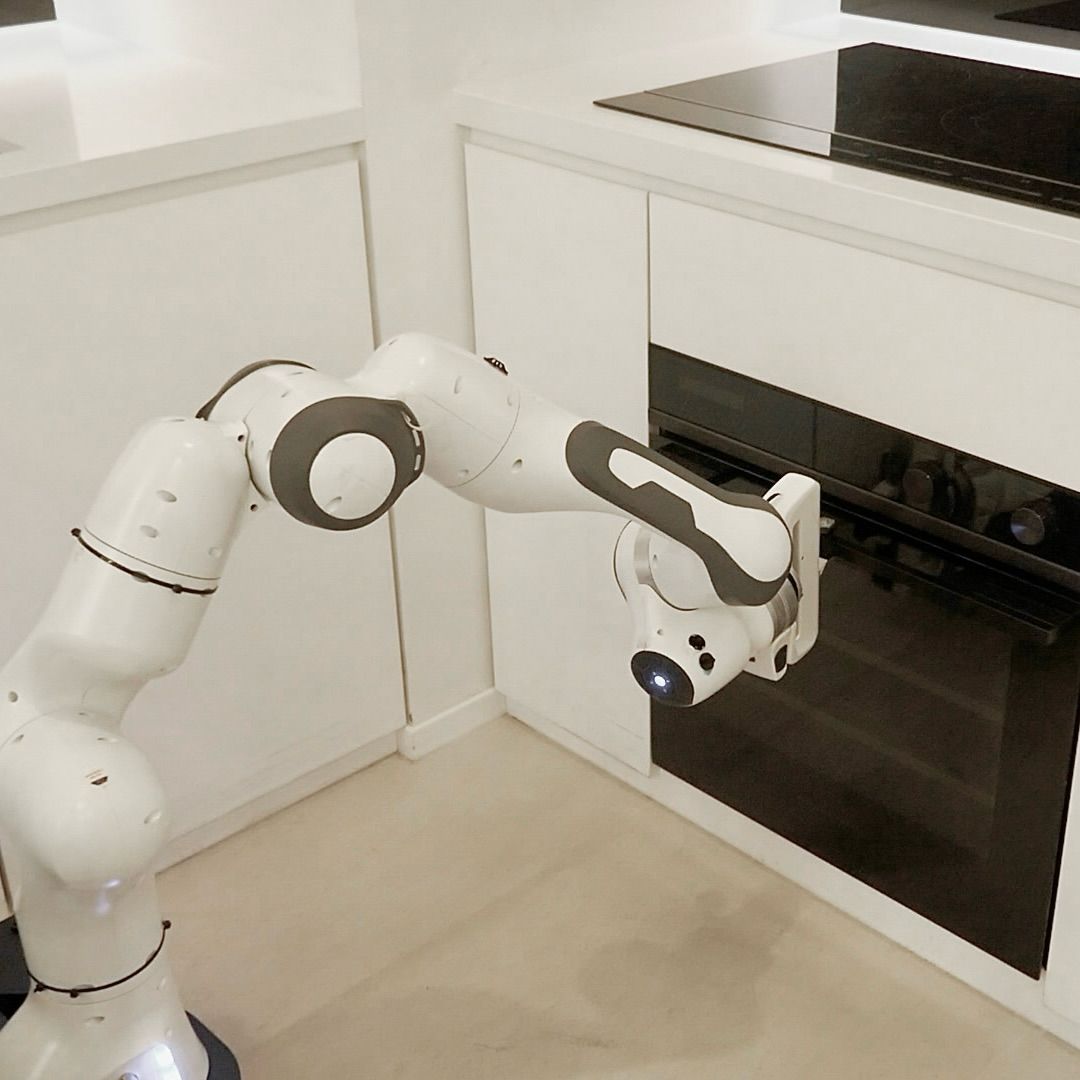} &
    \includegraphics[width=0.15\linewidth]{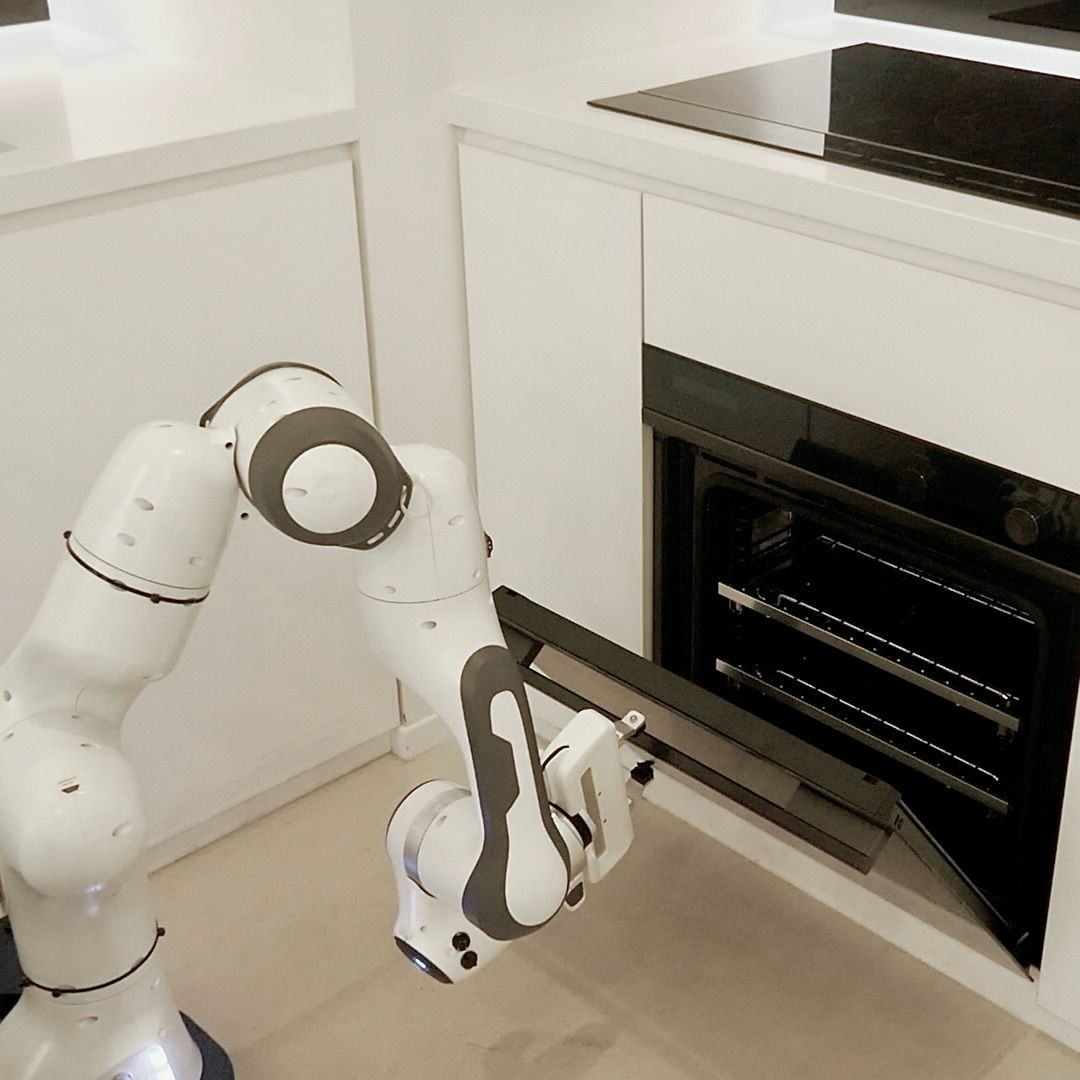} &
    \includegraphics[width=0.15\linewidth]{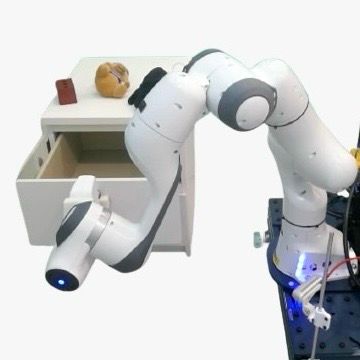} &
    \includegraphics[width=0.15\linewidth]{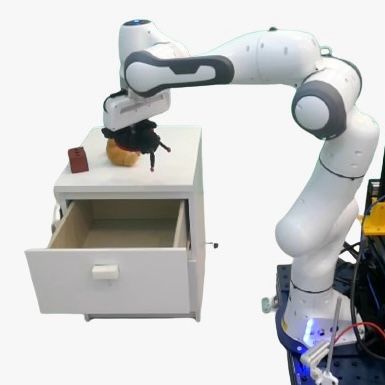} &
    \includegraphics[width=0.15\linewidth]{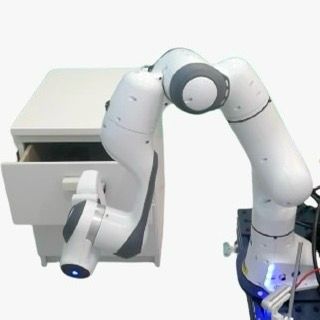} \\
    \multicolumn{2}{c}{\small (a) Open Oven} & \multicolumn{3}{c}{\small (b) Sort Objects into Drawer}
	\end{tabular}
	\centering
	\caption{\small Real-robot execution sequences. For both tasks, the robot needs to accurately predict the trajectories that understand the task context conditioned on languages. As appliances have high resistance force, a slight deviation from the expected trajectory would cause the robot to fail because of exceeding the joint torque limit.
 }\label{fig:real_robot}
	\label{fig:real_robot}
\end{figure*}
\subsection{Ablation Studies}
We perform ablation studies on the selected RLBench tasks to further understand the proposed low-level agent, \cdp{}. Since the high-level agent has been well-studied in prior works~\cite{shridhar2023perceiver}, we swap it with an expert and only focus on the performance of the low-level agents. We present the results in Tab.~\ref{tab:ablation}. 

\textit{Sampling-based motion planners might fail without understanding the task context.} As a sampling-based planner, RRT achieves a strong performance on simple tasks that only require goal information. However, for tasks that require a fine-grained trajectory, e.g., toilet seat up, RRT fails completely. As shown in Sect.~\ref{sect:vis}, we see that trajectories generated by RRT might easily violate the task constraints.\footnote{RLBench uses a hybrid motion planner of RRT and predefined linear paths by default. To reproduce the RRT result, we disable the linear path trajectories manually.} One could handcraft task-specific constraints, but it is not generalisable across tasks.

\textit{IK errors contribute to most of the failure cases of end-effector pose diffusion policy.} The \textit{Pose Diffusion} denotes learning a diffusion policy directly over the end-effector pose trajectories and generate robot controls by solving the inverse kinematics. We observe that although Pose Diffusion achieves strong performance on several tasks, e.g., open microwave, it suffers from an overall 24.55\% IK error rate. Specifically, most of the IK errors are caused by invalid quaternions and contribute to $75\%$ of its failure cases. In particular, the IK error rate increases as the control difficulty increases. This explains the importance of learning joint position trajectories, instead of end-effector poses.

\textit{Joint position diffusion is less accurate without the access to last joint position inpainting.} As in Sect.~\ref{sect:diffuser}, an end-effector pose will have multiple corresponding joint positions, and hence, it is infeasible for a joint position diffusion model to perform the last step inpainting. In our ablations, we show that it achieves a worse performance than \cdp{}, especially on challenging tasks, e.g., open oven.
    
\textit{3D information and the corresponding feature extraction module are critical to the performance of \cdp{}}. As mentioned earlier in Sect.~\ref{sect:diffuser}, \cdp{} uses a PointNet++ for point cloud feature extraction. For RKD-RGB, we discard the depth information and use a pretrained ResNet50 to extract the image features; for RKD-ResNet, we ablate using a ResNet to extract features from the RGB-D image. We observe that both achieve worse performance when compared to the original \cdp{}, which indicates that understanding the 3D environment is necessary for generalisable and accurate control. We believe there are alternative representations and leave it for future study.

\subsection{Real Robot Experiment}
We also conducted a real-world experiment on an opening oven task and a sorting objects into drawer task with a Franka Panda 7 DoF arm. We use 2 RealSense D415 cameras that captures the scene. For each sub-task we collect 10 demonstrations. Both tasks require the robot to accurately locate the target and control all its joints, especially the orientation of the wrist at every time step, otherwise, given the high resistance force of the oven, the arm will halt due to exceeding the joint torque limit. As a summary, HDP achieves 100\% success rate for the opening oven task and 94\% success rate for the sorting object into drawer task. Due to the nature of demo collection, we observe high variance in the demonstrated trajectories for the task. Intuitively, this leads to sub-optimal and highly diverse next-best pose predictions from the high-level agent, PerAct, some of which are out of distribution for \cdp{}. Interestingly, however, there appears to be minimal impact on \cdp{}, and the method is still capable of generalising to these unseen poses and generating accurate trajectories. Detailed results are in the appendix and are best viewed via the supplementary video.

\section{Conclusion}

We present \name{}, a hierarchical  agent for kinematics-aware robotic manipulation. \name{} factorises a policy: at the high-level, a task-planning agent predicts the next-best end-effector pose, and at the low-level, \cdp{} performs goal-conditioned prediction of a joint position trajectory that connects to the predicted next-best pose. To achieve both kinematics-awareness and high prediction accuracy, \cdp{} distills the accurate but less reliable end-effector pose trajectory to the joint position trajectory via differentiable kinematics. We show that \name{} achieves state-of-the-art performance on a set of challenging RLBench manipulation tasks. On a real robot, \name{} learns to solve both opening oven and sorting objects into drawer task. Although we have demonstrated some robustness of \cdp{} to out-of-distribution poses, the nature of behaviour cloning for longer-horizon tasks suggests that error accumulation could lead to significant distribution shifts and ultimate failure. Future works could explore improving the framework by designing more unified structures that minimises the compounding error. 

\newpage
{
    \small
    \bibliographystyle{ieeenat_fullname}

}
\newpage
\appendix

\section{Implementation Details}
\subsection{Perceiver-Actor}
The high-level Perceiver-Actor (PerAct) agent is a language-conditioned multi-task behaviour-cloning agent $\pi_\mathrm{high}(a\mid o, l)$, where $a_\mathrm{high} = (a_\pose, a_\mathrm{grip})$, $o$ consists of calibrated RGB-D multi-view images, and $l$ is a language description of the task. First, PerAct encodes the language description using the frozen pretrained CLIP~\cite{radford2021learning} language encoder. For the RGB-D images, PerAct computes the 3D position utilising camera intrinsics and extrinsics to obtain a 3D voxel grid representation of the current scene. Next, PerAct employs PerceiverIO to encode both language and voxel tokens with a fixed set of latent vectors. Finally, these vectors are decoded into a 3D action-value attention map. Specifically, we follow the official implementation and use $100^3$ voxels to represent the scene, and encode it 2048 fixed latent vectors with dimension 512. During training, we perform data augmentations by offsetting the voxels with random translations and rotations. During training, we minimise the following loss function
\begin{align}
    \label{eq:peract_loss} 
    \mathcal{L}_\mathrm{high} &= - \mathbb{E}_{k\sim\xi, \xi\sim \mathcal{D}}\left[ \log\pi_\mathrm{high}(a_\mathrm{demo}(k)\mid o, l)\right]\nonumber\\
    &= - \mathbb{E}_{k\sim\xi, \xi\sim \mathcal{D}}\left[ \log \pi_\mathrm{trans} + \log\pi_\mathrm{rot} + \log\pi_\mathrm{grip} \right]\nonumber\\
    \pi_\mathrm{trans} &= \mathrm{softmax}\left[Q_\mathrm{trans}((x, y, z)\mid o, l)\right]\nonumber\\
    \pi_\mathrm{rot} &= \mathrm{softmax}\left[Q_\mathrm{rot}((\psi_\mathrm{rot}, \theta_\mathrm{rot}, \phi_\mathrm{rot})\mid o, l)\right]\nonumber\\
    \pi_\mathrm{grip} &= \mathrm{softmax}\left[ Q_\mathrm{grip}(\omega\mid o, l) \right],
\end{align}
where $Q_\mathrm{trans}$, $Q_\mathrm{rot}$, and $Q_\mathrm{grip}$ are the discrete Q functions for the voxel-based translation policy $\pi_\mathrm{trans}$, the discrete rotation policy $\pi_\mathrm{rot}$, and the discrete binary gripper opening / closing policy $\pi_\mathrm{grip}$. Here, $(x, y, z)$ are the target voxel indices, $(\psi_\mathrm{rot}, \theta_\mathrm{rot}, \phi_\mathrm{rot})$ are the yaw, pitch, and roll parameters for a rotation, and $\omega$ is a binary value for gripper opening / closing. In particular, unlike the original PerAct with an additional action head $\pi_\mathrm{collision}$ to predict whether to ignore collision during RRT planning, we use \cdp{} as our low-level agent which handles collision checking automatically. Thus, we ignore $\pi_\mathrm{collision}$ for the high-level agent. During inference, PerAct directly takes the $\mathrm{argmax}$ of the discrete policy and reconstruct the continuous actions by indexing with the discrete indices. We directly take the hyper-parameters from the original PerAct~\cite{shridhar2023perceiver}.

\subsection{\cdp{}}
As discussed in the main text, in \cdp{}, we learn two separate diffusion models $\pi_\joint$ and $\pi_\pose$. As discussed in Sect.~\ref{sect:diffuser}, takes as the input the same conditions $C_\pose$, including the known start pose $a(0)_\pose^0$, the predicted next-best pose by the high-level agent $\hat{a}_\pose^0(T)$, the low-dimensional states $s$ of the robot, the gripper open amount $g$, and the point cloud of the environment $v$. Different from the voxel-based representation used in the high-level agent, we use point cloud for the low-level agent given the fact that low-level \cdp{} only requires understanding the 3D configuration of the environment, without predicting the action-values over the empty voxels. Besides, using voxels are  computationally more expensive than the point cloud. We use only the front camera. During inference, to perform conditional sampling, we use classifier-free guidance following~\citet{ho2022classifier}. 

For each vector input, we use a MLP with 2 hidden layers of sizes 128 and 512 with GELU activations~\cite{hendrycks2016gaussian}. For the point cloud, we use a standard PointNet++~\cite{qi2017pointnet++} as the encoder. We follow~\citet{janner2022planning} and use a Conv1D UNet~\cite{ronneberger2015u} as the temporal feature extractor. We use 3 repeated down-sampling residual blocks and 3 additional up-sampling residual blocks for the UNet. We train the goal-conditioned low-level setup in a multi-task setup, with 100 demonstrations per task. We optimise the networks for 100K steps with AdamW optimiser~\cite{loshchilov2017decoupled}. In addition, to compensate for the imperfect next-best pose predictions of the high-level agent, we perform data augmentations to the start pose and end pose by additionally taking $a(0 + k)$ and $a(T - k)$ as start / end pose along a sub-trajectory $\xi$, where $k\sim \mathcal{U}[0, 5]$. We determine the optimal hyper-parameters  through initial validation on three challenging RLBench tasks: open oven, open microwave, and open grill. These parameters are then applied across all tasks in both simulation and real-world settings. Our search covered the following hyper-parameters and training setups, with the selected parameters highlighted in bold: 1) Denoising steps: \textit{10, 50, \textbf{100}}; 2) Noise initialisation method: \textit{\textbf{normal distribution}} or \textit{uniform distribution}; 3) Approach: \textit{predicting the noise $\epsilon$} or \textit{\textbf{directly predicting the observation} $x_0$}. 

\section{Additional Results}
\begin{figure*}[t]
	\centering
	\begin{tabular}{ccccc}
    \includegraphics[width=0.18\linewidth]{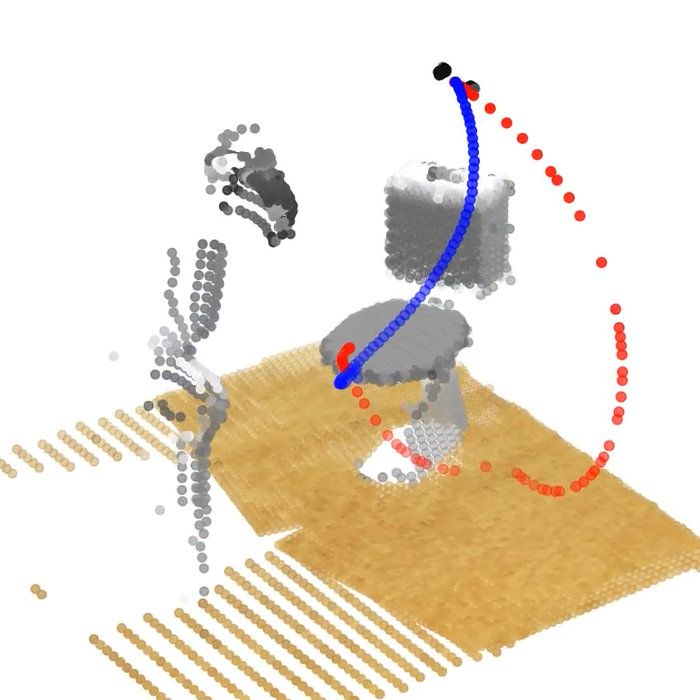} &
    \includegraphics[width=0.18\linewidth]{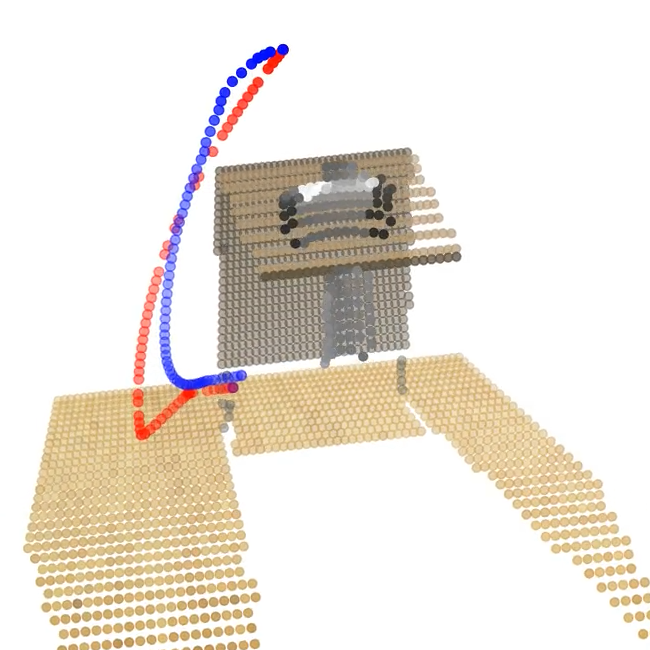} &
    \includegraphics[width=0.18\linewidth]{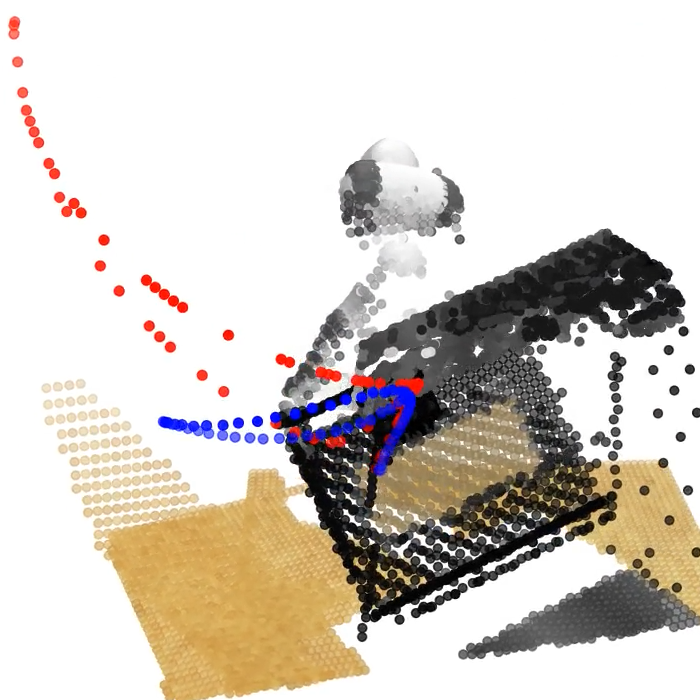} &
    \includegraphics[width=0.18\linewidth]{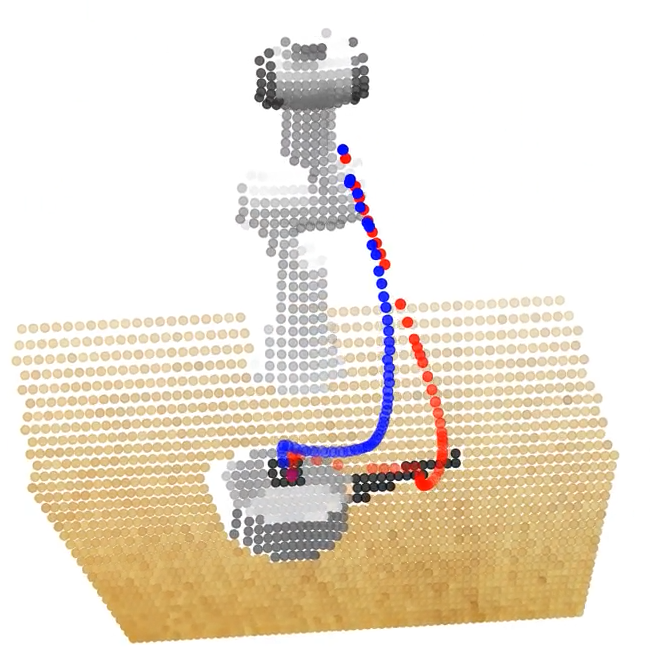} &
    \includegraphics[width=0.18\linewidth]{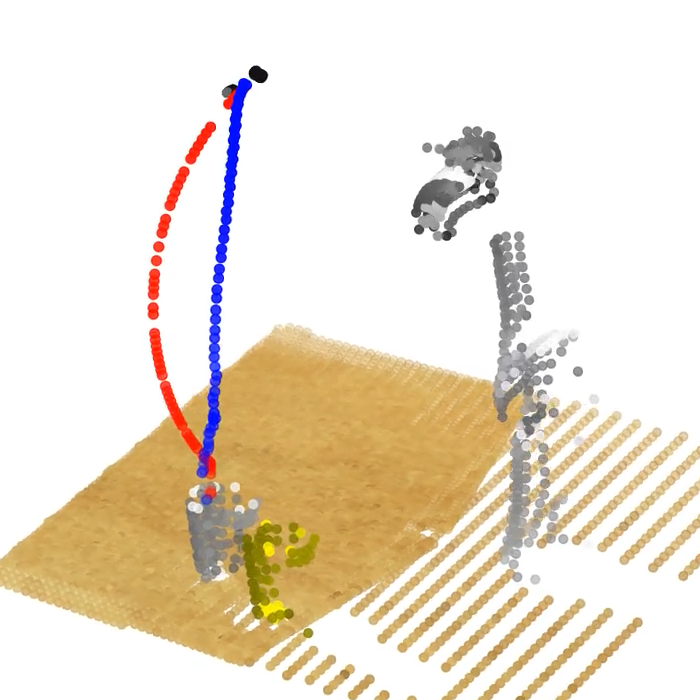}
	\end{tabular}
	\centering
	\caption{\small During training, we compute the rank of the trajectories, i.e., the optimality of the trajectory, to encourage the agent to differentiate quality of the sub-optimal trajectories generated by sample-based planners. During inference, we encourage \cdp{} to generate high-rank trajectories only, i.e., shorter trajectories. In this figure, \textcolor{red}{red} trajectories denote the ground-truth trajectories generated by planners, and \textcolor{blue}{blue} trajectories are generated by \cdp{}, which has shorter lengths while satisfying the kinematics-constraints.}\label{fig:traj_rank}
\end{figure*}

\subsection{Real-Robot Experiment Results}
\begin{table}[!htb]
\centering
\fontsize{7}{7}\selectfont
\caption{Real-robot experiment results for opening oven and sorting objects into drawer tasks.}
\begin{tabular}{ccccccccc}
\toprule
   & \multicolumn{2}{c}{Open Oven} & \multicolumn{5}{c}{Sort Objects into Drawer} & Overall     \\
   & reach          & open         & open   & beetle   & bear   & cube   & close  &             \\
   \midrule
\% & 100            & 100          & 100    & 85       & 100    & 85     & 100    & 95.71 \\
\bottomrule
\end{tabular}
\end{table}

\subsection{PerAct + Path Generation Baseline}

In the work~\cite{james2022lpr}, a set of trajectories are generated and ranked, reflecting the likelihood of success. Trajectories are generated using three methods including: sampled (linear + RRT), Bezier and learnt. Subsequently, over environment rollouts, an RL policy is trained to evaluate and rank these trajectories, with the highest being taken forward for execution. In the Behaviour Cloning setting, this direct approach is not applicable due to the absence of environment rollouts. Nevertheless, as part of the demonstration creation in RLBench~\cite{james2020rlbench}, the path generation method is recorded at each environment step and can be used as a training signal. To make use of this, an additional head is added to the PerAct backbone, which is trained to predict the optimal planning method -- either sampled or Bezier -- in addition to generating next-best-pose, gripper and ignore collision outputs. In this setting, the concept of collisions, as implemented in ~\cite{shridhar2023perceiver}, is crucial to enable reliable planning. Therefore, we adapt the loss function in equation \ref{eq:peract_loss}, incorperating the ignore collision policy, to include a path generation policy
\begin{align}
    \mathcal{L}_\mathrm{high} &= - \mathbb{E}_{k\sim\xi, \xi\sim \mathcal{D}}\left[ \log \pi_\mathrm{trans} 
    + \log\pi_\mathrm{rot} \right. \nonumber \\
    &\quad \left. + \log\pi_\mathrm{grip} + \log \pi_\mathrm{collision} 
    + \bold{\log\pi_\mathrm{path}}\right] \nonumber \\
    \pi_\mathrm{path} &= \mathrm{softmax}\left[ Q_\mathrm{path}(\lambda\mid o, l) \right],
\end{align}
where $Q_\mathrm{path}$ is the discrete Q function for the voxel-based path planning policy $\pi_\mathrm{path}$.
Here, $\lambda$ is a discrete selection for optimal planner method. During inference, if the model determines that a sampled approach is optimal, it initially attempts a linear path, followed by RRT planning if necessary. In the case of the Bezier method, the model samples a set of random curvature parameters, executing the first successful configuration, if any.

\subsection{Trajectory Ranking}
As discussed in Sect.~\ref{sect:diffuser}, we include an additional conditional variable, trajectory rank, into \cdp{}. We define trajectory rank as $r_\xi = \frac{d_\mathrm{Euclidean}}{d_\mathrm{travel}}$, where $d_\mathrm{Euclidean}$ is the Euclidean distance between the start and end pose and $d_\mathrm{travel}$ is the travelled distance between the start and end pose. Intuitively, an optimal trajectory, ignoring the kinematics constraint of the robot, should have $r_\xi = 1$. We provide additional visualisations to analyse the effect of trajectory ranking in Fig.~\ref{fig:traj_rank}, where the red curves denote the ground-truth trajectory, and blue ones are the trajectories sampled from \cdp{}. We observe that the \cdp{} is capable to generate shorter trajectories while trained with sub-optimal planner demonstrations.

\end{document}